%% file: main.tex
\documentclass{article}

\PassOptionsToPackage{numbers,compress}{natbib}
\usepackage[dblblindworkshop,final]{neurips_2026}
\workshoptitle{NeurIPS 2026 Workshops}

\makeatletter
\renewcommand{\@trackname}{%
  40th Conference on Neural Information Processing Systems
  (NeurIPS 2026) Workshops.%
}
\makeatother

\usepackage[utf8]{inputenc}
\usepackage[T1]{fontenc}
\usepackage{hyperref}
\usepackage{url}
\usepackage{booktabs}
\usepackage{subcaption}
\usepackage{amsmath}
\usepackage{amsfonts}
\usepackage{nicefrac}
\usepackage{microtype}
\usepackage{xcolor}
\usepackage{graphicx}
\usepackage{array}
\usepackage{tabularx}
\usepackage{caption}
\usepackage{float}
\usepackage{longtable}
\usepackage{enumitem}

\definecolor{detblue}{RGB}{224,239,250}
\definecolor{judgmentorange}{RGB}{253,237,216}
\definecolor{verifygreen}{RGB}{226,242,231}
\newcommand{\method}[1]{\textsc{#1}}

\title{Evaluating and improving AI agents for sequencing protocol understanding}

\author{
Chi-Lam Poon$^{1,2,}$\thanks{Corresponding author: \texttt{poonc2@mskcc.org}} \quad
Michael Goneos$^{2,3}$ \quad
Zezheng Song$^{4}$ \quad
Xi Chen$^{5}$ \\[0.5em]
\small
$^{1}$Memorial Sloan Kettering Cancer Center \quad
$^{2}$Weill Cornell Medicine \\
\small
$^{3}$NASA Ames Research Center \quad
$^{4}$University of Maryland, College Park \\
\small
$^{5}$Southern University of Science and Technology
}

\begin{document}

\maketitle

\begin{abstract}
Existing scientific-agent benchmarks rarely evaluate whether agents
can reconstruct the exact molecular structures and generation workflows
specified by sequencing protocols, a capability critical for analyzing
customized sequencing assays. We introduce LibStructBench, a
document-grounded benchmark for reconstructing sequencing-library
structures and molecular workflows. Our experiments show that frontier
agents often produce approximately correct reconstructions but remain
unreliable at exact reconstruction, particularly in molecular-state
representation and workflow connectivity.
Switching the underlying language model produced larger average performance
differences than switching agent harnesses, with a statistically
significant difference only for molecular-state reconstruction.
Additional knowledge improved oligo-family recovery in a small pilot
without improving mean molecular-workflow scores.
Autonomous accumulation of procedural rules and cross-protocol examples
showed non-monotonic validation performance and no mean held-out
molecular-workflow improvement in the tested trajectory.
These findings highlight the need for more controlled
and verifiable approaches to scientific reasoning about
molecular processes in complex experimental protocols.
\end{abstract}

\section{Introduction}
Correct analysis of sequencing data depends on knowing the molecular structure
of the library being sequenced. Adapter sequences, cell and sample barcodes,
unique molecular identifiers (UMIs), their positions and strand orientation
determine how raw sequencing reads should be extracted, trimmed, demultiplexed,
and quantified. This is particularly important for customized or newly
developed single-cell assays, where barcode and UMI layouts may differ
substantially from standard commercial protocols. Machine-readable
specifications such as \method{seqspec} \citep{booeshaghi2024seqspec} 
demonstrate that explicit library descriptions can support standardized 
preprocessing across genomics assays. However, such approaches assume that the
underlying library structure has already been established.

For new or customized assays, establishing this structure can itself require
substantial manual interpretation of the experimental design, followed by
custom code for cell and molecule recovery and assessment of assay sensitivity and
specificity. Moreover, diagnosing a failed sequencing run is even harder: 
it requires not only the final read structure but also an understanding of all molecular intermediates and experimental operation steps. 
AI agents based on frontier large language models (LLMs) could
potentially automate this process by reconstructing the library, generating
assay-specific preprocessing logic, inspecting raw sequencing data, and tracing
quality-control failures back to candidate experimental steps. Such capability
could substantially accelerate the development and debugging of new sequencing
technologies.

Sequencing protocols distribute relevant evidence across text, supplementary 
tables and files, and figures. An agent must therefore identify which components 
are relevant, assign them to the correct molecular roles and orientations, 
and track how intermediate molecules are transformed into the final library. 
This makes sequencing-protocol reconstruction a stringent test of
evidence-based domain-specific reasoning rather than simply document retrieval.
In this work, we introduce LibStructBench to measure AI agent's ability to extract
and reconstruct the molecular logic of sequencing protocols from their source
documents. The benchmark decomposes this problem into three linked tasks that model
the physical construction of sequencing libraries (Figure~\ref{fig:benchmark-overview}): 
Task~1 reconstructs the final sequencing-library structure; 
Task~2 recovers workflow-linked oligo families and
their roles; and Task~3 reconstructs library generation as a graph of
molecular states and transitions. Human curated ground-truth records were
audited against protocol evidence.

Our experiments show that frontier LLMs often reconstruct
final-library sequences approximately but remain unreliable
at exact reconstruction. For molecular-workflow reconstruction,
no native system dominated all metrics, and model-associated
differences were most pronounced for molecular-state
reconstruction. Failure analysis identified frequent
discrepancies in molecular-state assembly, workflow
connectivity, and missing or unsupported components.
The broad baseline findings remained largely robust to
alternative scoring assumptions, while repeated executions
revealed variability at the individual-protocol level.
Protocol-name-only controls further demonstrated substantial
reconstruction capability without supplied documents,
highlighting the importance of verifying protocol-specific
information against primary evidence.
Neither additional domain knowledge nor cumulative learning
consistently improved molecular-workflow reconstruction
in the tested settings. These observations
motivate the development of agents that can explicitly
ground and verify molecular structures and transformations.

\section{Benchmark construction and validation}

LibStructBench was built from \texttt{scg\_lib\_structs}, an open resource of
manually reconstructed sequencing-library primer and adapter elements
(hereafter, oligos) and step-by-step library-generation procedures for
single-cell genomics methods
(\url{https://teichlab.github.io/scg_lib_structs/}). The resource spans diverse
assay classes, including gene expression, chromatin accessibility, DNA
methylation, genomic DNA, and multi-omic protocols. We selected more than 
70 curated commercial and academic sequencing protocols
with diverse library architectures as the initial legacy records for
LibStructBench.

For each protocol, we assembled an agent-visible evidence bundle containing
available experimental protocol or technical documentation, primary journal article,
supplementary files, oligo tables, and molecular figures. These bundles were
frozen and content-addressed before evaluation and defined the evidence boundary
for reconstruction. Agents could inspect and process the supplied files but
were not permitted to search the web or retrieve external protocol-specific
information. Final ground-truth records were produced only after the agent-audit
and human-adjudication procedure described below.

The benchmark datasets and implementation are publicly available:
\begingroup
\urlstyle{same}
\begin{itemize}[
    leftmargin=1.3em,
    itemsep=1pt,
    topsep=2pt,
    parsep=0pt
]
\raggedright

\item \textbf{Protocol ground truth:}
\url{https://huggingface.co/datasets/sequencing}


\item \textbf{Benchmark implementation:}
\url{https://github.com/seqmachines/libstruct-bench}

\item \textbf{Workflow visualization and agent traces:}
\url{https://www.acolytics.com/proofread}

\end{itemize}
\endgroup

The ground-truth release manifests retain recorded source URLs, document
versions where available, and content hashes. These manifests do not grant
redistribution rights for the underlying documents, which retain their
original licensing restrictions.

\subsection{Benchmark tasks and output representations}

\begin{figure*}[t]
\centering
\includegraphics[width=\textwidth]{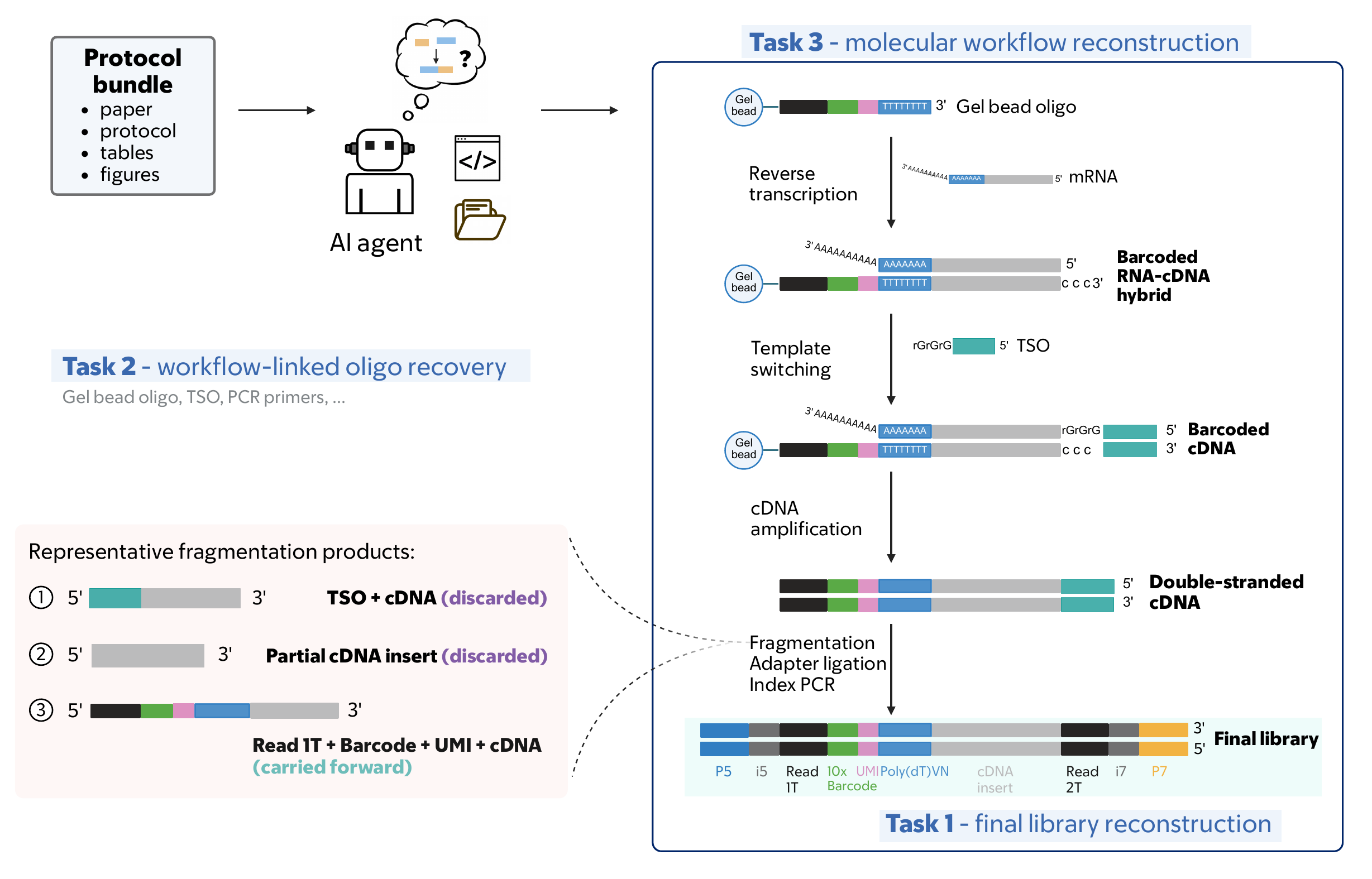}
\caption{%
\textbf{LibStructBench overview.}
Given a protocol bundle containing primary source evidence, an AI agent
reconstructs three linked representations of sequencing-library construction,
corresponding to Tasks~1-3.
The schematic illustrates these tasks using a simplified workflow for
10x Chromium 3$'$ Gene Expression v4 \citep{tenx2024gemxv4}.
}
\label{fig:benchmark-overview}
\end{figure*}

\paragraph{Task 1: final sequencing-library reconstruction.}
The final library structure is a key piece of information for raw
sequencing-data processing and sequencing run quality assessment. Given a protocol bundle, the
agent must reconstruct the canonical $5'\rightarrow3'$ structure of each
defined final library. Fixed nucleotide segments and typed variable regions are
normalized before scoring. Reconstruction accuracy is evaluated using normalized 
sequence similarity based on Levenshtein distance (edit distance),
\begin{equation}
s(x,y)=1-\frac{d_{\mathrm{Lev}}(x,y)}{\max(|x|,|y|)}.
\label{eq:task1-score}
\end{equation}

\paragraph{Task 2: workflow-linked oligo recovery.}
Oligos are the physical reagents that include adapters, priming sites, barcodes, 
and other technical elements into a sequencing library. Extracting them therefore tests whether the reconstructed molecular workflow is grounded in the correct building blocks, and the ground truth of this task is mainly the "Adapter and primer sequences" section in \texttt{scg\_lib\_structs}. The agent returns one structured record per oligo family, including its sequence, name and alias, role, molecular kind, orientation, ordered components, and modifications. Ordered components
are used for assembled, double-stranded, or hairpin oligos that cannot be
represented faithfully as a single flat sequence.

Required ground-truth oligo families are those linked to Task~3
with source-recoverable sequences. Agent-extracted oligos with
compatible sequence scaffolds and metadata are collapsed into
predicted families and matched one-to-one to required families,
with eligible optional or non-recoverable matches neutralized. 
For each extraction-ground-truth pair, we compute a weighted similarity
\begin{equation}
w_{ij}
=
0.65\,s_{\mathrm{seq}}
+0.15\,s_{\mathrm{mod}}
+0.10\,s_{\mathrm{kind}}
+0.05\,s_{\mathrm{orient}}
+0.05\,s_{\mathrm{role}}.
\label{eq:task2-sim}
\end{equation}

The weights summarize sequence, modification, molecular kind,
orientation, and role similarity, and are renormalized over
supported components.
If \(\mathcal{M}^{*}\) denotes the resulting one-to-one matching, the primary Task~2 score is
\begin{equation}
F1_{\mathrm{oligo}}
=
\frac{
2\sum_{(i,j)\in\mathcal{M}^{*}} w_{ij}
}{
N_{\mathrm{out}}^{\mathrm{eff}} + N_{\mathrm{req}}
}.
\label{eq:oligo-f1}
\end{equation}

Here, \(\mathcal{R}\) denotes required, source-recoverable ground-truth
families, \(\mathcal{P}\) all collapsed predicted families, and
\(\mathcal{Z}\subseteq\mathcal{P}\) predictions neutralized by eligible
optional or non-recoverable matches. Neutralization uses exact matches
or a correspondence threshold of 0.50 for used non-recoverable families,
while preserving required matches. Thus,
\(N_{\mathrm{req}}=|\mathcal{R}|\) and
\(N_{\mathrm{out}}^{\mathrm{eff}}=|\mathcal{P}\setminus\mathcal{Z}|\),
counting both matched and unmatched non-neutralized predictions.
When \(N_{\mathrm{req}}=0\), F1 is 1 if
\(N_{\mathrm{out}}^{\mathrm{eff}}=0\), and 0 otherwise.
These protocols are excluded from primary Task~2 macro-F1.

\paragraph{Task 3: molecular library-generation workflow.}
Task~3 reconstructs the ``Step-by-step library generation'' section in
\texttt{scg\_lib\_structs} as a directed state-transition graph. Shared
upstream steps are represented once, with modality-specific or alternative
routes branching toward their terminal states. Each state records the physical
molecular strands and their \(5'\!\rightarrow\!3'\) architecture, while each
transition records the input states, molecular operation, oligos,
products, and whether those products are carried forward or discarded. For example, 
reverse transcription (operation) converts an RNA template
(substrate) using an RT primer (oligo) into an RNA--cDNA intermediate (product). And in the fragmentation example shown in
Figure~\ref{fig:benchmark-overview},
fragmentation generates multiple product classes from double-stranded 
cDNA (substrate), with the barcode-bearing
cDNA fragment carried forward and internal or TSO-end fragments discarded.

A transition is represented as
$\tau =
\left(
S_{\mathrm{in}},
O,
Q,
S_{\mathrm{out}},
S_{\mathrm{carried}},
S_{\mathrm{discarded}}
\right),
$
where \(S_{\mathrm{in}}\) and \(S_{\mathrm{out}}\) are the input and output
molecular states (e.g. RNA and RNA-cDNA intermediate), \(O\) is the molecular operation, and \(Q\) is the set of
linked Task~2 oligo families. Agent-reconstructed and ground-truth workflows are 
first matched using their terminal modalities, states, and transitions. 
Within each matched workflow, states and transitions
are aligned through global partial one-to-one matching,
with candidate pairs below a context-adjusted assignment
score of \(0.25\) left unmatched.
For each matched transition pair, scientific similarity is
\begin{equation}
s^{\tau}_{ij}
=
0.30s_{\mathrm{operation}}
+
0.15s_{\mathrm{substrates}}
+
0.20s_{\mathrm{products}}
+
0.15s_{\mathrm{disposition}}
+
0.20s_{\mathrm{oligos}},
\label{eq:task3-trans-sim}
\end{equation}
which compares the operation, aligned input and product states,
carried-versus-discarded disposition, and transition-linked oligos.

For entity type \(X\in\{\tau,S\}\), representing transitions or states, let
\(\mathcal{M}^{*}_{X}\) denote the matched pairs,
\(N_{\mathrm{out}}^{X,\mathrm{eff}}\) the number of scored agent-output
entities, and \(N_{\mathrm{gt}}^{X,\mathrm{score}}\) the number of scorable
ground-truth entities. Transition and state F1 are computed using the same
soft-F1 form:
\begin{equation}
F1_X
=
\frac{
2\sum_{(i,j)\in\mathcal{M}^{*}_{X}} s^{X}_{ij}
}{
N_{\mathrm{out}}^{X,\mathrm{eff}}
+
N_{\mathrm{gt}}^{X,\mathrm{score}}
},
\qquad X\in\{\tau,S\}.
\label{eq:task3-soft-f1}
\end{equation}
Matched entities receive fractional credit, unmatched scored outputs reduce precision,
and missing ground-truth entities reduce recall. For states,
\(s^{S}_{ij}\) weights reference-strand sequence \(0.50\), molecular
architecture \(0.15\), ordered strand and segment structure \(0.20\), and
pairing or discontinuities \(0.15\).

To evaluate graph topology separately, we compare typed substrate,
carried-product, and discarded-product edges between aligned states. Let
\(m_E\) denote the number of matching typed edges. Using the same effective
output and scorable ground-truth counts, edge F1 is
\begin{equation}
F1_{\mathrm{edge}}
=
\frac{
2m_E
}{
N_{\mathrm{out}}^{E,\mathrm{eff}}
+
N_{\mathrm{gt}}^{E,\mathrm{score}}
}.
\label{eq:task3-edge-f1}
\end{equation}

\subsection{Ground-truth construction, audit, and adjudication}
Human-curated records in \texttt{scg\_lib\_structs} may contain
reconstruction errors, while the underlying protocol documents may
themselves be ambiguous or internally inconsistent. 
We therefore audited every candidate benchmark record
before evaluation. For each protocol, the legacy curation was first converted
into the canonical Task~1--3 ground-truth schema. An AI agent then
compared this candidate record against the frozen source bundle and generated
proposed corrections for human review.

Codex implemented the deterministic audit framework, including source
manifests, schemas, Task~1--3 validation, review checkpoints,
and application of approved changes. Claude Code performed the protocol-level
audit, checking each candidate against the protocol sources and flagging
unsupported claims, ambiguities, protocol-version conflicts, and proposed
corrections. A human expert adjudicated every substantive finding by accepting, revising,
or rejecting the proposed change and directly reviewed the oligos, molecular
states, and transitions against the primary sources. Only human-approved changes
were applied to the benchmark record (Figure~\ref{fig:supp-audit-loop}).
In the summarized audit, 54 of 69 confirmed findings (78.3\%) were
categorized as legacy-curation errors (Figure~\ref{fig:supp-audit-errors}).
This statistic describes error causes rather than the origin
of accepted corrections, which could not be reliably determined
from the available adjudication records. Future ground-truth audit updates 
and corrections will be documented
at \url{https://acolytics.com/proofread}.

\section{Experiments and results}
\subsection{Interface and PDF-conversion controls}
\paragraph{Direct LLM API and coding-agent comparison.}
We first assessed final-library reconstruction accuracy and whether native
coding-agent interfaces improved performance over direct LLM APIs.
Three models were evaluated across
\href{https://huggingface.co/datasets/sequencing/scg-protocols-v1-groundtruth}
{20 sequencing protocols}:
Claude Opus 5, GPT-5.6 Sol, and Gemini 3.7 Flash, using their direct
APIs and corresponding native agent interfaces (Claude Code, Codex, and
Antigravity CLI) with high reasoning effort via Harbor \citep{harbor2026}.
Direct APIs received protocol PDFs without tools, whereas
agents could use native file, shell, and document-processing tools.
Excel sheets were excluded in inputs here because direct APIs 
do not accept this format.

\begin{figure}[h]
  \centering
  \makebox[\linewidth][c]{%
    \begin{minipage}[c]{0.68\linewidth}
      \centering
      \includegraphics[width=\linewidth]{
        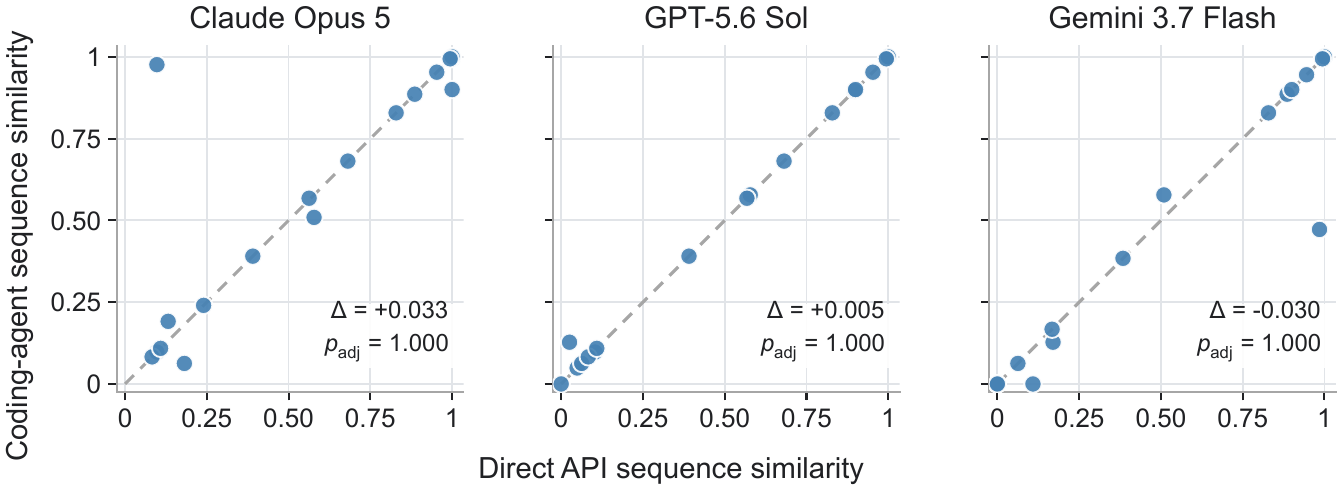
      }
    \end{minipage}%
    \hspace{0.02\linewidth}%
    \begin{minipage}[c]{0.28\linewidth}
      \centering
      \textbf{\small Paired Task 1 performance}
      \par\vspace{0.25em}
    
      {\small
      \setlength{\tabcolsep}{2pt}
      \renewcommand{\arraystretch}{1.3}
    
      \begin{tabular}{@{}lcc@{}}
        \toprule
        Model &
        \shortstack{Mean sim.\\API/Agent} &
        \shortstack{Exact/20\\API/Agent} \\
        \midrule
        Claude & .591/.623 & 5/3 \\
        GPT    & .561/.566 & 4/4 \\
        Gemini & .597/.567 & 4/4 \\
        \bottomrule
      \end{tabular}
      }
    \end{minipage}%
  }
  \caption{%
    \textbf{Task 1 performance for direct APIs versus native
    coding agents.}
    Each point represents one sequencing protocol completed
    by both interfaces.
    $\Delta$ denotes the mean paired similarity difference
    (agent minus API), and $p_{\mathrm{adj}}$ denotes the adjusted 
    $p$-value for two-sided paired Wilcoxon signed-rank test.
    The inset table reports mean normalized sequence similarity
    and exact reconstruction counts for each interface.
  }
  \label{fig:task1-api-agent}
\end{figure}

Mean paired similarity differences (agent minus API) were +0.033,
+0.005, and -0.030 for Claude, GPT, and Gemini, respectively.
None was statistically significant in two-sided paired Wilcoxon
signed-rank tests with Bonferroni correction across the three
model comparisons ($p_{\mathrm{Bonf}}=1.000$ for all;
Figure~\ref{fig:task1-api-agent}). Exact final-library reconstruction 
remained uncommon, with only 3-5 of 20 protocols (15-25\%) 
reconstructed exactly across configurations;
API/agent win counts were 4/3 (13 ties), 1/1 (18 ties), and
3/1 (16 ties), respectively (Figure~\ref{fig:task1-api-agent}).
Therefore, native coding agents showed no statistically detectable
improvement in Task~1 accuracy.

As final-library similarity alone cannot characterize errors
in molecular components and workflow reconstruction, we further
evaluated Tasks~2 and 3 using coding agents, whose tool-calling
and structured-execution capabilities support more complex
evaluations and improvement.

\paragraph{Text-layer nucleotide sequences are reliably recovered from PDFs.} 
Since sequence reconstruction errors could arise from technical PDF text conversion 
rather than downstream protocol reasoning, we tested sequence recovery 
across five representative protocols containing 80 human-verified,  
source-visible nucleotide strings. Curated human text, 
PyMuPDF \citep{pymupdf}, pypdf \citep{pypdf}, 
Docling without OCR \citep{livathinos2025docling}, and 
MinerU OCR \citep{wang2024mineru} were compared, 
and a fixed Codex independently recovered sequence strings. 
Each ground-truth sequence \(y_j\) was scored by its best match among the 
recovered candidates, \(\max_i s(x_i,y_j)\), using the normalized Levenshtein 
sequence similarity defined in Eq.~\eqref{eq:task1-score}. 

Text from human, PyMuPDF, pypdf, and Docling recovered all 80 sequences exactly (Figure~\ref{fig:parser-control}), but MinerU OCR recovered only 71 of 80 
exact sequences. The OCR errors included corruption of 
repetitive character strings such as "GGGGGG". These results suggest that 
text-layer nucleotide extraction alone is unlikely to explain the 
observed reconstruction errors. However, this does not assess 
sequences embedded in figures or the interpretation of graphical molecular structures.

\subsection{Baseline molecular-workflow reconstruction}
\begin{figure*}[h]
  \centering
  \captionsetup[subfigure]{skip=1pt}

  \begin{subfigure}[t]{0.35\textwidth}
    \centering
    \includegraphics[width=\linewidth]{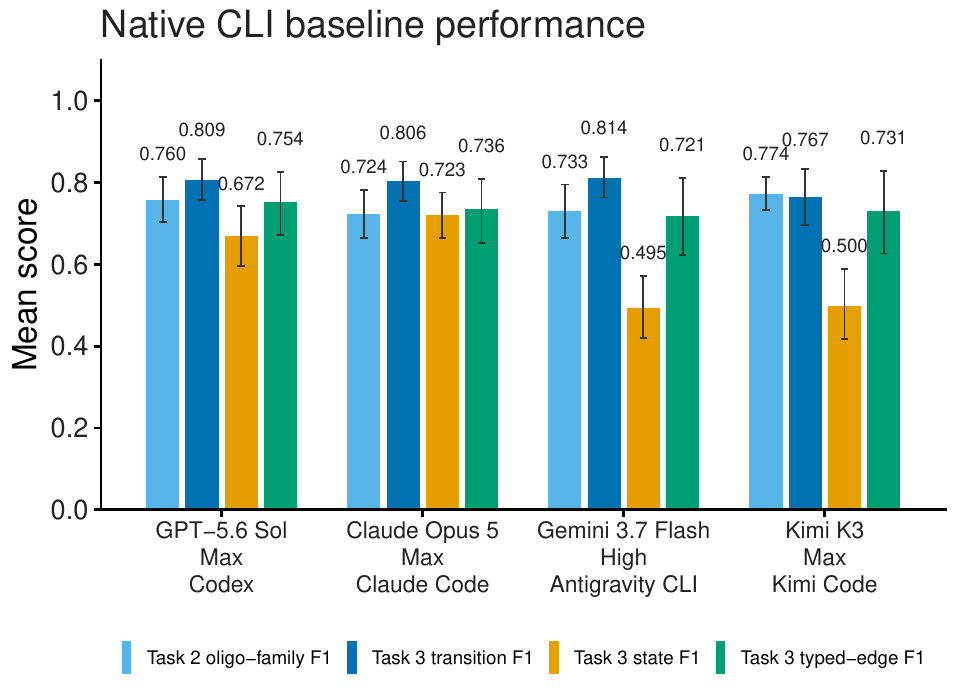}
    \caption{}
    \label{fig:baseline-native}
  \end{subfigure}
  \hfill
  \begin{subfigure}[t]{0.62\textwidth}
    \centering
    \includegraphics[width=\linewidth]{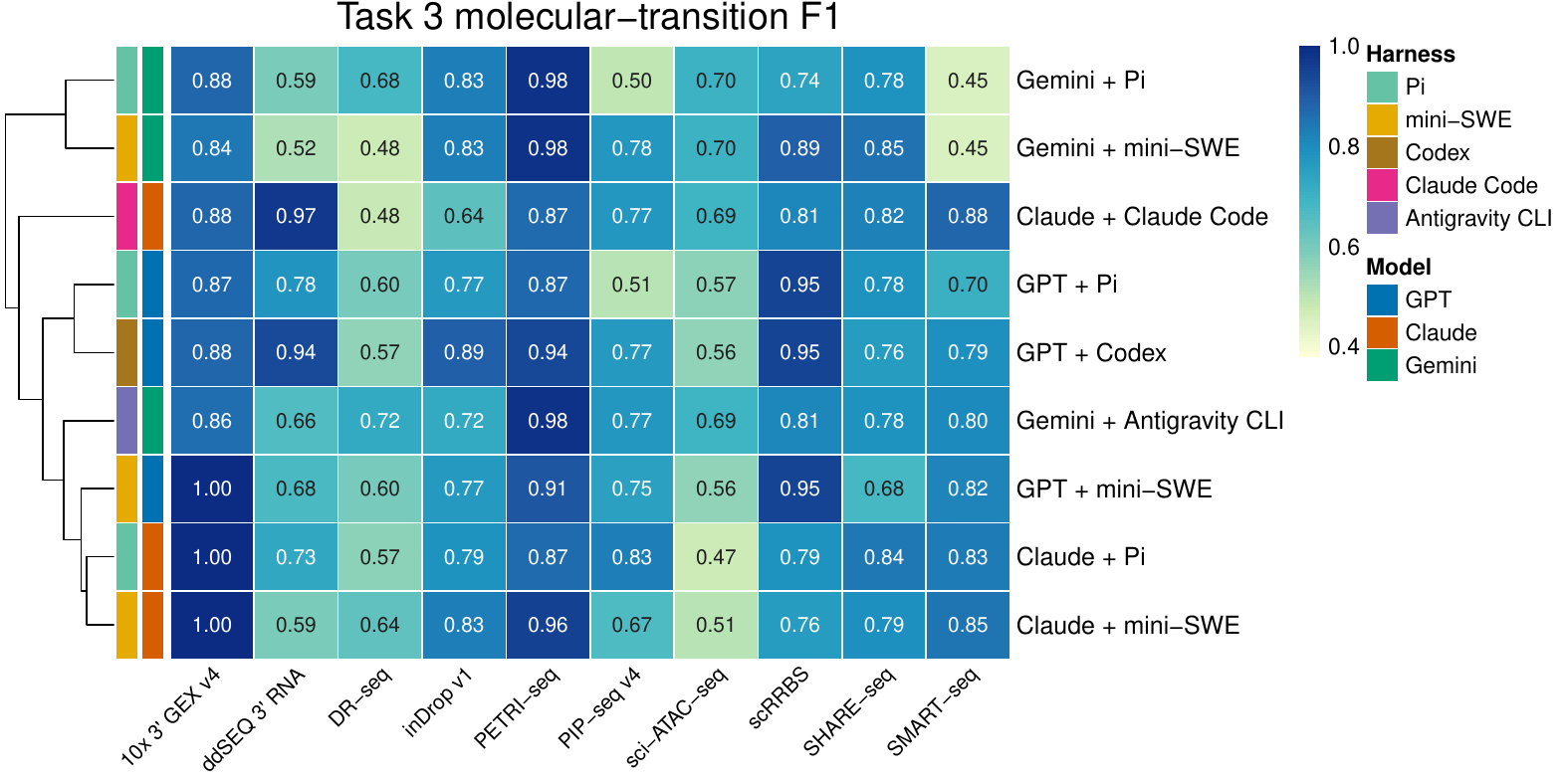}
    \caption{}
    \label{fig:baseline-transition}
  \end{subfigure}
  \caption{%
    \textbf{Baseline agent comparison.}
    \textbf{(a)}~Mean Task 2 oligo-family and Task 3 transition, state, and
    typed-edge F1 for native CLI systems.
    \textbf{(b)}~Task 3 molecular-transition F1 across 10 common protocols
    and nine model-harness systems, clustered by protocol-level performance.
  }
  \label{fig:baseline-comparison}
\end{figure*}

\paragraph{Baseline performance across models and harnesses.}
To assess how reliably coding agents could reconstruct the underlying
molecular workflow and components, we benchmarked four native
coding-agent systems on the same 20-protocol panel: GPT-5.6 Sol Max with Codex,
Claude Opus 5 Max with Claude Code, Gemini 3.7 Flash High with Antigravity CLI,
and Kimi K3 Max with Kimi Code. Each system received the same frozen protocol
bundles, task instructions, output schema, and deterministic verifier similarly 
through Harbor \citep{harbor2026}. 
For Task~2, ddSEQ was excluded from the macro-average because it
contained no required, source-recoverable oligo families
($N_{\mathrm{req}}=0$). Mean Task~2 oligo-family F1 ranged from 0.724 to 0.774 with required recoverable oligo families.
Task~3 molecular-transition, molecular-state, and typed-edge F1
ranged from 0.767--0.814, 0.495--0.723, and 0.721--0.754,
respectively, across all 20 protocols (Figure~\ref{fig:baseline-native}).
No single system dominated all metrics. Kimi obtained the highest
Task~2 F1, Gemini the highest transition F1, Claude the highest
state F1, and GPT the highest typed-edge F1.
Molecular-state F1 showed the largest between-system variation.

We then asked whether the agent harness altered performance for the same
underlying model. GPT, Claude, and Gemini were additionally paired with 
Pi \citep{pi2026} and mini-SWE-agent \citep{yang2024sweagent}, yielding 
nine model-harness systems when combined with their native interfaces. 
These systems were compared on 10 protocols common to all configurations, 
drawn from the 20-protocol baseline panel. Since molecular-transition F1 jointly evaluates the operation, 
substrates, products, linked oligos, and product disposition, 
we used it as the primary Task~3 metric for this comparison. 
Mean molecular-transition F1 ranged from 0.714 to 0.805 across 
the nine model-harness systems
(Figure~\ref{fig:baseline-transition}). Hierarchical clustering did not 
clearly separate systems by model or harness, although molecular-state 
performance showed more pronounced model-associated patterns.
(Supplementary Figure~\ref{fig:supp-complementary-heatmaps}).

Estimated costs varied across systems
(Figure~\ref{fig:supp-cost-by-system})
and were not monotonically related to reconstruction performance
(Figure~\ref{fig:supp-cost-performance}).
To quantify model and harness effects, we performed a
protocol-paired analysis across three models and two shared
harnesses (Pi and mini-SWE-agent).
Model-switch differences exceeded harness-switch differences
on average across all four metrics, although only molecular-state
F1 remained statistically significant after Bonferroni correction
($\Delta=0.079$, pointwise 95\% CI [0.038, 0.124],
$p_{\mathrm{Bonf}}=0.016$;
Supplementary Figure~\ref{fig:supp-model-harness}).

\paragraph{Variability across repeated executions.}
As our baseline comparisons used a single execution
per model-harness system and protocol, we assessed how
aggregate performance and individual-protocol scores varied
across repeated executions of the same agent configuration.
We evaluated GPT-Codex across three executions per protocol
on the 20-protocol baseline panel.
Averaging the three executions changed the original
oligo-family, molecular-transition, molecular-state, and
typed-edge macro-F1 scores by $-0.005$, $-0.014$,
$-0.012$, and $-0.024$, respectively.
Despite these relatively small aggregate differences,
within-protocol F1 ranges exceeded 0.10 for 5/19
informative oligo protocols and 9/20, 8/20, and
12/20 protocols for the three Task~3 metrics,
respectively (Appendix~\ref{app:execution-variability}).
Although individual protocols showed execution-to-execution
variability, the aggregate performance estimates remained
broadly consistent across the three executions, with substantial
room for improvement in molecular-workflow reconstruction.
Given the computational cost of repeated evaluations across
multiple models and harnesses, we retained single-execution
comparisons for the broader benchmark.

\paragraph{Sensitivity to scoring assumptions.}
Similarity weights and matching thresholds
may influence reported scores and system rankings,
we therefore recomputed matching and scoring across 640 configurations
(Appendix~\ref{app:metric-sensitivity}).
The leading native systems for oligo-family, molecular-state,
and typed-edge F1 remained unchanged. Typed-edge F1 was also more sensitive to
matching thresholds than transition or state F1.
Overall, the broad baseline findings were robust to the
tested scoring assumptions, although absolute scores
and close system rankings warrant cautious interpretation.

\paragraph{Reconstruction without protocol source bundles.}
Many sequencing protocols and \texttt{scg\_lib\_structs} are publicly available
and may have appeared in LLM training data, 
we further evaluated protocol-name-only controls on the
20-protocol panel (Appendix~\ref{app:name-only-control}).
In a separate Task~1 API experiment, name-only prompting
increased mean sequence similarity for all three models
but yielded fewer exact reconstructions. Only Claude's
similarity improvement remained significant after
Bonferroni correction ($p_{\mathrm{adj}}=0.0215$).
For native Tasks~2-3, name-only prompting yielded
lower mean F1 across all three systems and four metrics,
although none of the 12 paired comparisons remained
significant after Bonferroni correction.
These results indicate substantial reconstruction
ability without supplied evidence but do not establish
training-data memorization or that source bundles
are unnecessary.

\paragraph{Failure modes.}
To characterize where agent reconstructions disagreed with the ground truth, we
applied deterministic error analysis to 100 outputs from ten systems across ten
common protocols. For each output, the verifier recorded whether at least one
discrepancy occurred in each of six predefined categories
(Table~\ref{tab:baseline-error-terms}). Molecular-state assembly and 
workflow-topology/disposition discrepancies were identified in all 100 outputs
(Figure~\ref{fig:supp-baseline-errors-system}). Missing recoverable information
occurred in 88/100 outputs, unsupported additions in 77/100, operation or target
mismatches in 24/100, and strand or orientation errors in only 8/100. 
Exploratory clustering of system-level discrepancy profiles
tended to group configurations by underlying LLM
(Figure~\ref{fig:supp-baseline-errors-system}), while
protocol-level profiles showed variation in missing information,
unsupported additions, and operation or target mismatches
(Figure~\ref{fig:supp-baseline-errors-protocol}).
These categories are nonexclusive, as molecular-state
alignment errors can propagate into topology discrepancies;
their frequencies therefore describe disagreement with the
canonical reconstruction rather than independent failure mechanisms.

\subsection{Improving molecular-workflow reconstruction}
Accurate protocol understanding is important for downstream sequencing quality
control, where errors in library structure and oligo definitions can
directly affect read parsing and quantitative analyses. Baseline results showed
that none of the native model-harness systems achieved consistently accurate 
reconstruction, we next explored whether supplying additional domain knowledge 
or accumulating reviewed experience could improve agent performance. 

\begin{figure}[h]
  \centering
  \captionsetup[subfigure]{skip=1pt}

  \begin{subfigure}[t]{0.45\textwidth}
    \centering
    \includegraphics[width=\linewidth]
      {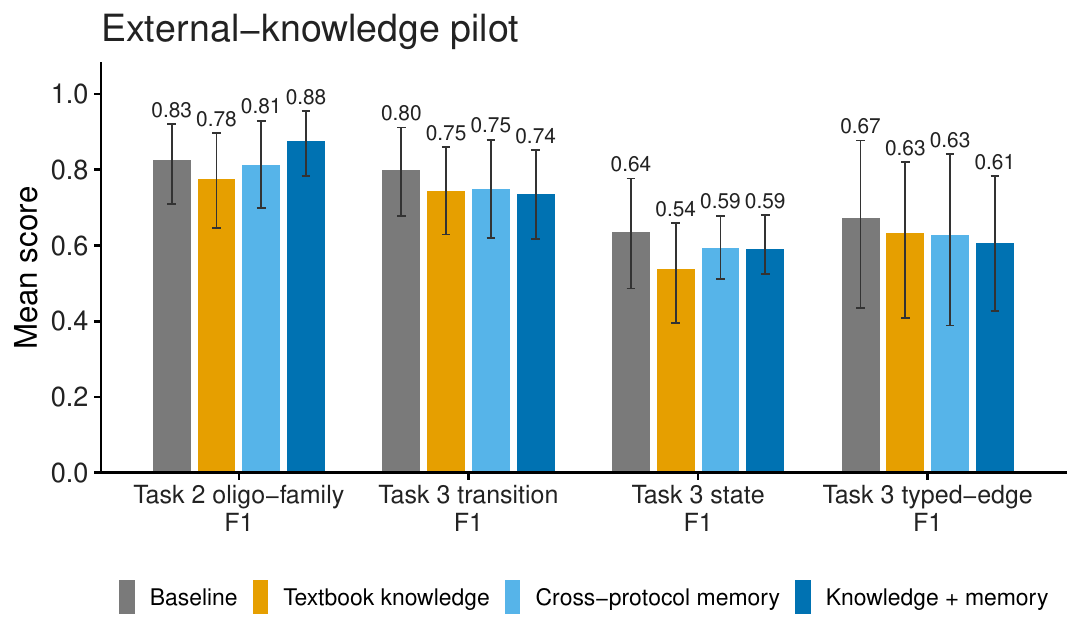}
    \caption{}
    \label{fig:knowledge-pilot}
  \end{subfigure}
  \hfill
  \begin{subfigure}[t]{0.49\textwidth}
    \centering
    \includegraphics[width=\linewidth]
      {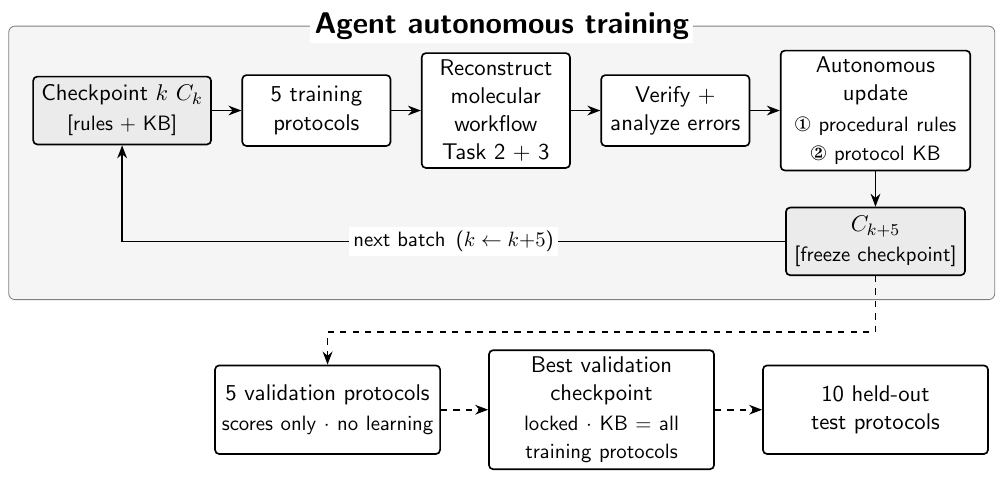}
    \caption{}
    \label{fig:autonomous-training}
  \end{subfigure}

  \vspace{0.5em}

  \begin{subfigure}[t]{0.45\textwidth}
    \centering
    \includegraphics[width=\linewidth]
      {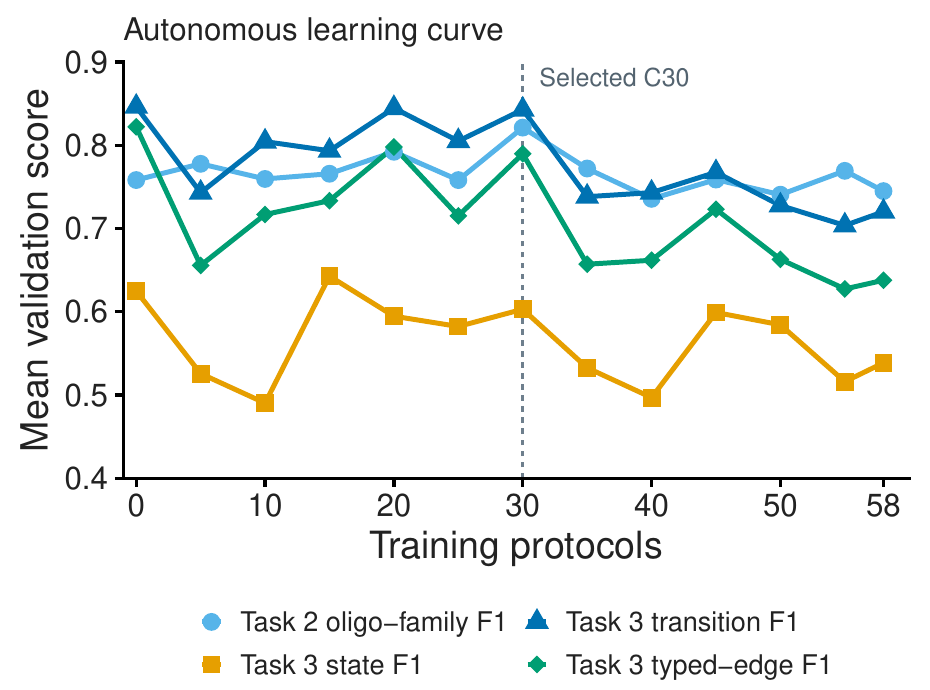}
    \caption{}
    \label{fig:agent-learning-curve}
  \end{subfigure}
  \hfill
  \begin{subfigure}[t]{0.45\textwidth}
    \centering
    \includegraphics[width=\linewidth]
      {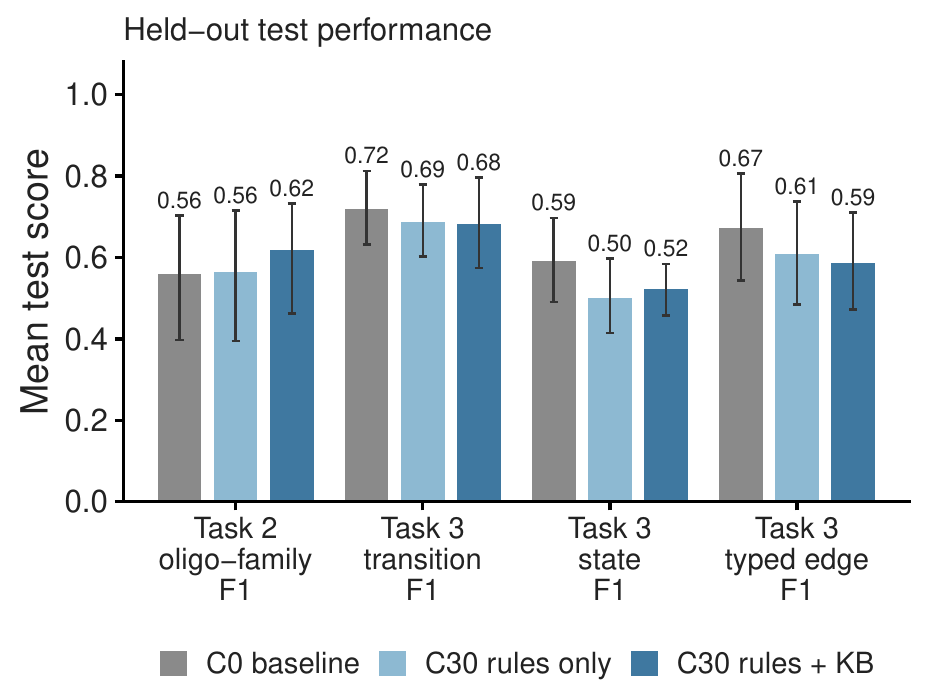}
    \caption{}
    \label{fig:autonomous-test}
  \end{subfigure}

  \caption{%
    \textbf{Improving molecular-workflow reconstruction.}
    \textbf{(a)}~Mean Task~2 and Task~3 scores under the baseline,
    general-knowledge, cross-protocol memory, and combined conditions.
    \textbf{(b)}~Autonomous cumulative-improvement framework.
    \textbf{(c)}~Validation performance across autonomous-training
    checkpoints.
    \textbf{(d)}~Held-out test performance of baseline and
    the validation-selected checkpoint with and without
    knowledge base (KB) supplement. For panels (a) and (c), 
    Task~2 includes four informative
    validation protocols and Task~3 includes five; the two
    experiments use different validation panels.
    Panel (d) includes nine informative held-out protocols for
    Task~2 and ten protocols for Task~3.%
  }
  \label{fig:improvement}
\end{figure}

\paragraph{External knowledge had task-dependent effects in this pilot.}
GPT-5.6 Sol Max with Codex was used to examine whether 
external domain knowledge 
or cross-protocol examples could improve Tasks~2 and~3 performance on a fixed
five-protocol validation panel (sci-ATAC-seq, scRRBS, SMART-seq, SHARE-seq, ddSEQ). Each protocol was tested under four paired
conditions: 1) the protocol bundle alone (baseline); 2) a general
molecular-method primer derived from textbooks and review articles 
(Table~\ref{tab:general-primer-sources}); 
3) ground-truth records from five separate donor protocols, reformatted as example
agent outputs and supplied as cross-protocol exemplar memory 
(Table~\ref{tab:pilot-protocol-pairing}); and 4) both knowledge sources. 
The baseline achieved mean oligo-family, molecular-transition, 
molecular-state, and typed-edge F1 scores of 0.826, 0.800, 0.636, 
and 0.673, respectively (Figure~\ref{fig:knowledge-pilot}). 
Combining general knowledge with cross-protocol memory increased
oligo-family F1 to 0.877, a paired improvement of 0.050
(pointwise 95\% protocol-bootstrap CI [0.022, 0.079]).
General knowledge alone and memory alone achieved oligo-family F1
scores of 0.777 and 0.814, respectively.
All three augmented conditions had lower Task~3 mean scores than
baseline, although their paired confidence intervals included zero.
These results show a task-dependent response in the saved outputs:
combined knowledge improved informative-protocol oligo recovery,
while no augmentation exceeded baseline Task~3 point estimates.

\paragraph{Autonomous cumulative learning.}
We tested whether an agent could improve molecular-workflow reconstruction by
accumulating experience across previously verified protocols. The 73 usable
protocols were split into 58 training, 5 validation, and 10 held-out test
protocols. Validation panel includes CEL-seq, chATAC-seq, ddSEQ, 
HyDrop-RNA, scRRBS. Training proceeded in batches of five protocols. We denote the
initial agent as checkpoint \(C_0\) (baseline), 
and each subsequent checkpoint \(C_k\) as
the agent state after learning from \(k\) training protocols. After each batch,
the agent autonomously updated two components: 1) a protocol-neutral procedural
rule set derived from observed agent errors and 2) a knowledge base of oligos
and molecular graphs derived from completed training ground truth. These
components were frozen as the next checkpoint, and each checkpoint was evaluated
on the same five validation protocols without using validation outputs for
learning (Figure~\ref{fig:autonomous-training}).

Validation performance was non-monotonic across checkpoints. Using the aggregate
validation criterion, \(C_{30}\) was selected, whereas performance declined at
later checkpoints, with substantial changes appearing by \(C_{30}\)-\(C_{40}\).
The final checkpoint \(C_{58}\) scored below \(C_0\) on all four validation
metrics (Figure~\ref{fig:agent-learning-curve}). We then evaluated the 
selected \(C_{30}\) procedural rules on the 10 held-out
protocols, both alone and supplemented with the accumulated training
knowledge base. Relative to \(C_0\), the \(C_{30}\) rules alone did not improve
any Task~3 metric. 
Adding the knowledge base increased the held-out oligo-family
F1 point estimate from 0.560 to 0.621
(paired difference 0.061; pointwise 95\% protocol-bootstrap CI
[-0.024, 0.159]).
Transition, state, and typed-edge F1 means remained below baseline
(0.684 vs.\ 0.721, 0.523 vs.\ 0.592, and 0.588 vs.\ 0.674,
respectively; Figure~\ref{fig:autonomous-test}).
Thus, this single learning trajectory showed non-monotonic
validation performance and no held-out Task~3 point-estimate gain.
These findings are conditional on the saved outputs and canonical
graph representation; they do not establish that cumulative
learning generally fails to improve molecular understanding.

\section{Discussion}
Our results highlight a gap between recovering sequencing-library
components and faithfully reconstructing molecular workflows.
Agents often identified experimental operations but struggled
with intermediate molecular states and workflow connectivity,
explaining relatively high transition F1 despite lower state F1.
Task~2 and Task~3 scores measure agreement with canonical
references rather than independently validated biological
correctness or downstream usability. The broad findings were
robust to tested scoring assumptions, while repeated GPT-Codex
executions revealed protocol-level variability despite similar
aggregate performance.

\paragraph{Model and harness effects.}
No native model-harness system consistently outperformed
the others across all reconstruction metrics. In our paired
comparison across three models and two shared harnesses,
switching the underlying language model produced larger
average performance differences than switching harnesses,
with a statistically significant difference only for
molecular-state reconstruction. This suggests that
representing intermediate molecular architectures is
particularly sensitive to the underlying model in the
tested configurations.

\paragraph{Additional knowledge and autonomous learning had task-dependent effects.}
Combining general biological knowledge with cross-protocol
examples improved oligo-family recovery in our small pilot
without improving mean molecular-workflow scores.
Similarly, accumulated knowledge increased the held-out
oligo-family F1 point estimate, with uncertainty spanning
zero, but did not improve Task~3 mean performance.
These findings suggest that recovering additional molecular
components does not necessarily translate into correctly
integrating them into a molecular workflow.

One possible explanation is that models already possess
substantial general sequencing knowledge, as suggested
by the name-only controls. Additional context may therefore
offer limited benefit while complicating the selection
and integration of protocol-specific evidence.
Moreover, cumulative learning showed non-monotonic
validation performance, with some learned rules associated
with omitted molecular intermediates or altered workflow
connectivity. Some discrepancies may also reflect
differences in representation granularity, as chemically
valid intermediate steps can be penalized by canonical
graph matching. Furthermore, the verifier may penalize chemically valid
reconstructions when their molecular-state or workflow
representations differ from the canonical ground truth.

\paragraph{Future improvement should emphasize controlled and verifiable learning.}
Our document-grounded setting reflects the intended application
to new or unpublished sequencing protocols, for which
protocol-specific information may not be publicly available.
We therefore use publicly documented protocols from
\texttt{scg\_lib\_structs} as an auditable benchmark,
while restricting agents to frozen source bundles rather
than external retrieval.

The name-only controls suggest that agents can leverage
prior knowledge to generate candidate reconstructions,
which could then be verified against protocol-specific evidence.
Future agents could first propose a provisional molecular
workflow, then extract evidence from supplied documents to
verify, revise, or reject individual components and transitions.
Model Context Protocol (MCP)-accessible tools could support
document parsing, oligo validation, molecular-state construction,
and graph-consistency checking, while cross-protocol memory
could enable iterative improvement.
Learned rules should be reversible and validated against
previously solved protocols before retention.
Future work should compare this prior-guided approach with
direct document-grounded reconstruction and accommodate
biologically equivalent workflow representations.

\section*{Acknowledgments}
We thank Ruslan Soldatov (Memorial Sloan Kettering Cancer Center)
and Xinming Tu (Phylo Bio) for valuable discussions and guidance.
We also acknowledge the Google Cloud Research Credits Program for
providing Google Cloud and Gemini API credits, and the
MIT--Anthropic AI for Science program and Claude hackathon
for providing Claude usage and API credits.

\clearpage
\bibliographystyle{plainnat}
\bibliography{references}

\clearpage
\appendix
\renewcommand{\thefigure}{S\arabic{figure}}
\setcounter{figure}{0}
\renewcommand{\thetable}{S\arabic{table}}
\setcounter{table}{0}

\section{Related work}
\paragraph{Sequencing-library specifications.}
Explicit descriptions of sequencing-library structure are important for
reproducible upstream preprocessing and interpretation of single-cell sequencing data.
Community resources such as \texttt{scg\_lib\_structs}
(\url{https://teichlab.github.io/scg_lib_structs/}) manually curate primers,
adapters, barcodes, final-library structures, and library-generation steps
across single-cell protocols. \method{seqspec} \citep{booeshaghi2024seqspec} formalizes this idea with a machine-readable specification for genomics assays, representing ordered library regions, technical sequences, and their relationships to sequencing reads. 
This enables more standardized preprocessing, assay comparison, and reanalysis 
across various sequencing technologies. LibStructBench addresses a complementary 
problem: rather than assuming that a structured assay description is already available, 
it evaluates whether an AI agent can autonomously reconstruct such information 
accurately from heterogeneous sequencing-protocol documents.

\paragraph{Biomedical agents and benchmarks.}
Recent agentic systems increasingly combine language-model reasoning with
scientific tools, retrieval, and structured execution environments. Co-Scientist \citep{gottweis2026coscientist} uses a multi-agent
architecture to generate, critique, and refine scientific hypotheses, and
Robin \citep{ghareeb2026robin} combines literature-search and data-analysis
agents in an iterative lab-in-the-loop discovery workflow. \method{BIOMNI}
\citep{huang2026biomni} generalizes this idea to a broad biomedical environment
spanning database retrieval, code execution, and other domain tools.

On the evaluation side, LAB-Bench \citep{laurent2024labbench},
ScienceAgentBench \citep{chen2025scienceagentbench}, BiomniBench
\citep{qu2026biomnibench}, and DrugDiscoveryBench
\citep{akyurek2026drugdiscoverybench} evaluate increasingly realistic biological and
scientific tasks, from literature and sequence reasoning to executable
multi-step workflows. LibStructBench differs by focusing on a
narrow but exact reconstruction problem: recovering sequencing-library
components and the molecular workflow from primary protocol files.

\paragraph{Deterministic verification and iterative improvement.}
Recent agent systems increasingly combine generative reasoning with
deterministic feedback and persistent adaptation. VirBench
\citep{nasri2026virbench} shows that coupling scientific agents to a
deterministic programmatic retrieval layer substantially improves the
reliability of viral-sequence retrieval. SAGE \citep{wang2026sage} develops a
self-improving agent with a persistent skill library, in which skills learned
from earlier tasks accumulate. DeepVerifier \citep{wan2026deepverifier} 
instead uses rubric-based verification to generate
feedback for iterative refinement at inference time. Together, these studies
motivate closed-loop agent systems where deterministic tools and prior
experience are converted into reusable capabilities. Our work follows
this direction by testing whether self-improvement loops that learn from
discrepancies on reviewed training protocols can improve molecular-workflow 
reconstruction of unseen sequencing protocols.

\clearpage
\section{Human-agent ground-truth audit and adjudication}
\begin{figure*}[h]
  \centering
  \begin{subfigure}[t]{0.67\textwidth}
    \centering
    \includegraphics[width=\linewidth]{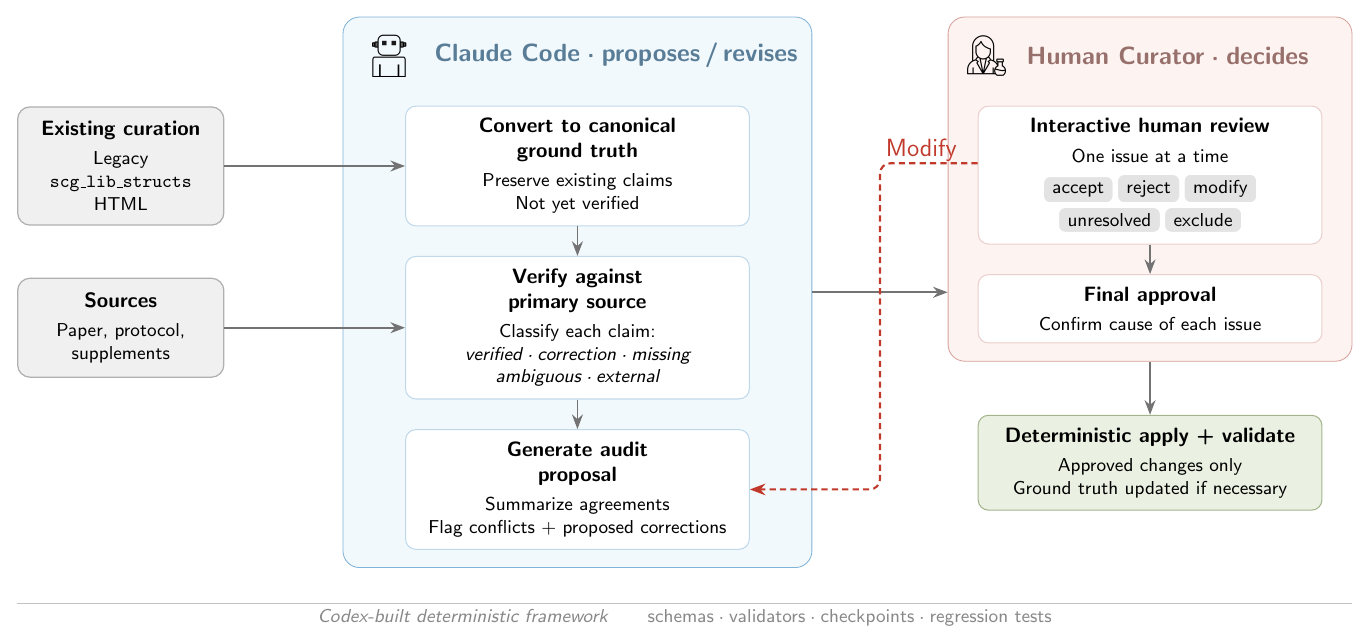}
    \caption{}
    \label{fig:supp-audit-loop}
  \end{subfigure}
  \hfill
  \begin{subfigure}[t]{0.33\textwidth}
    \centering
    \includegraphics[width=\linewidth]{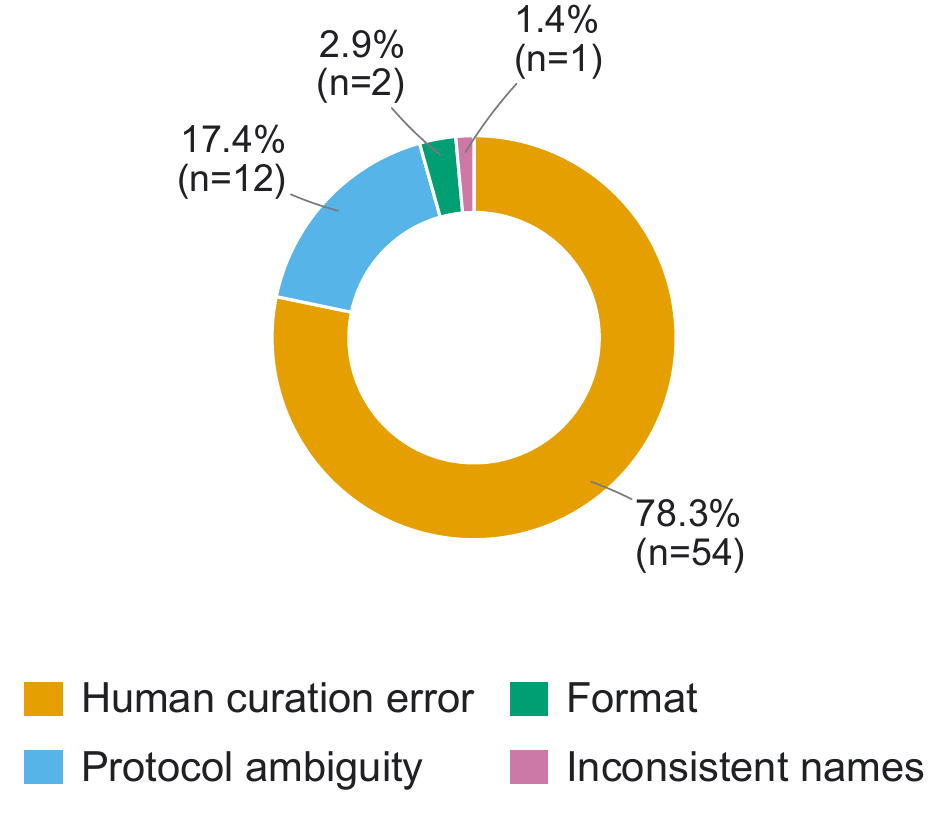}
    \caption{}
    \label{fig:supp-audit-errors}
  \end{subfigure}

  \caption{%
    \textbf{Human-agent ground-truth audit and adjudication.}
    \textbf{(a)}~Existing human curation is converted to the canonical
    representation and verified against primary evidence.
    Claude Code proposes corrections, a human curator adjudicates
    substantive findings, and only approved decisions are applied
    by deterministic code.
    \textbf{(b)}~Distribution of confirmed audit findings of errors from 73 usable 
    protocols collected in \texttt{scg\_lib\_structs}.%
  }
  \label{fig:supp-audit}
\end{figure*}

\paragraph{Representative errors.} \

\textbf{Human curation error (dsciATAC-seq).}
The legacy record labeled the 6-nt sequence between the s5 adapter and mosaic
end as a UMI. The Methods and oligo table identify it as one of 96 Tn5 well
barcodes, so the placeholder was corrected from \texttt{[UMI:6]} to
\texttt{[TN5\_INDEX:6]} in Tasks 1 and 3.

\textbf{Protocol ambiguity (scDamID).}
The supplement prints a 57-nt P5 primer beginning \texttt{ATG}, whereas the
final-library structure in the same protocol begins \texttt{AATG} and requires
the 58-nt primer. Direct PDF re-extraction confirmed that the source itself
omits the leading A; we retained the 58-nt form and recorded the conflict.

\section{PDF-parser-control detail}
\begin{figure}[h]
  \centering
  \begin{minipage}[c]{0.58\linewidth}
    \centering
    \includegraphics[width=\linewidth]{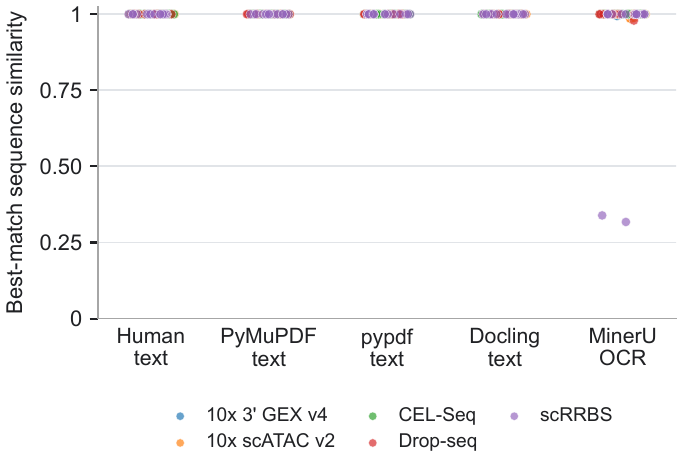}
  \end{minipage}\hfill
  \begin{minipage}[c]{0.38\linewidth}
    \centering
    \small
    \setlength{\tabcolsep}{3pt}
    \renewcommand{\arraystretch}{1.12}
    \begin{tabular}{lrr}
      \toprule
      Input & Exact & Mean sim. \\
      \midrule
      Human text & 80/80 & 1.000 \\
      PyMuPDF text & 80/80 & 1.000 \\
      pypdf text & 80/80 & 1.000 \\
      Docling text & 80/80 & 1.000 \\
      MinerU OCR & 71/80 & 0.982 \\
      \bottomrule
    \end{tabular}
  \end{minipage}
  \caption{%
    \textbf{PDF-parser-control sequence recovery.}
    Sequence recovery was evaluated across 80 verified nucleotide strings 
    from 5 protocols. For each reference sequence, we report the highest 
    normalized Levenshtein similarity to any sequence extracted. 
    The table summarizes exact recovery and mean similarity.%
  }
  \label{fig:parser-control}
\end{figure}

\section{Variability across repeated agent executions}
\label{app:execution-variability}
We assessed execution variability for GPT-5.6 Sol Max
with Codex by comparing $n=1$ versus $n=3$ executions
per protocol across the 20-protocol baseline panel
(60 executions total).
The $n=3$ analysis included the original execution.
All outputs were evaluated using the same scoring rules
and approved ground-truth references as the main experiments.
For protocol $p$, metric $m$, and execution $r$,
let $F_{prm}$ denote the corresponding F1 score.
Within-protocol variability was measured as
$R_{pm}=\max_r F_{prm}-\min_r F_{prm}$
across the three executions.
We summarized these ranges across protocols using
the median and interquartile range (linearly interpolated
empirical quartiles), and counted protocols with $R_{pm}>0.10$.
Execution-specific protocol-macro F1 was calculated as
$\bar{F}_{rm}=\frac{1}{P_m}\sum_{p=1}^{P_m}F_{prm}$,
and the three-execution mean as
$\bar{F}_{\cdot m}=\frac{1}{3}\sum_{r=1}^{3}\bar{F}_{rm}$,
where $P_m$ denotes the number of informative protocols.
Task~2 excludes ddSEQ because $N_{\mathrm{req}}=0$
($P_m=19$), whereas Task~3 includes all 20 protocols
($P_m=20$).
\begin{figure}[htbp]
\centering
\begin{minipage}[t]{0.42\textwidth}
  \centering
  \includegraphics[
    width=\linewidth,
    height=0.25\textheight,
    keepaspectratio
  ]{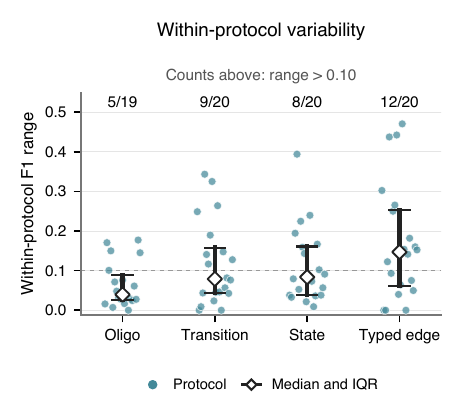}
  \par\vspace{2pt}
  (a)
\end{minipage}
\hspace{0.025\textwidth}
\begin{minipage}[t]{0.42\textwidth}
  \centering
  \includegraphics[
    width=\linewidth,
    height=0.25\textheight,
    keepaspectratio
  ]{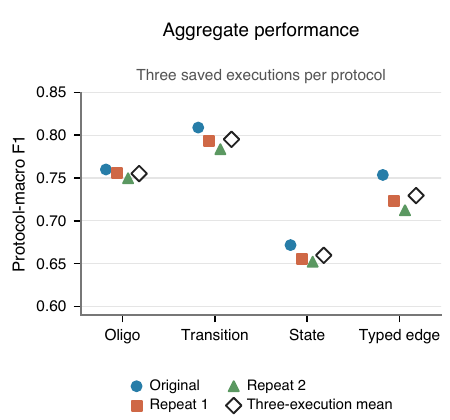}
  \par\vspace{2pt}
  (b)
\end{minipage}
\caption{%
\textbf{GPT-Codex variability across three executions per protocol.}
\textbf{(a)} Within-protocol F1 ranges.
Points represent protocols; diamonds and intervals indicate
medians and interquartile ranges. The dashed line marks
a range of 0.10, with annotations indicating counts above
this threshold.
\textbf{(b)} Protocol-macro F1 for each execution and
their three-execution mean.
}
\label{fig:supp-execution-variability}
\end{figure}

\begin{table}[htbp]
\centering
\small
\setlength{\tabcolsep}{5pt}
\renewcommand{\arraystretch}{1.08}

\begin{tabular}{@{}lrrrrr@{}}
\toprule
Metric & Original & 3-run mean & $\Delta$ & Max. range & $R>0.10$ \\
\midrule
Oligo family & 0.760 & 0.755 & $-0.005$ & 0.177 & 5/19 \\
Transition   & 0.809 & 0.795 & $-0.014$ & 0.343 & 9/20 \\
State        & 0.672 & 0.660 & $-0.012$ & 0.394 & 8/20 \\
Typed edge   & 0.754 & 0.730 & $-0.024$ & 0.471 & 12/20 \\
\bottomrule
\end{tabular}
\vspace{0.5em}

\caption{\textbf{Observed variability across three GPT-Codex executions per protocol.}
Macro-F1 averages 19 informative protocols for oligo recovery and 20 for each
Task~3 metric.}
\label{tab:execution-variability}
\end{table}

The three-execution macro-F1 estimates differed from
the original scores by only 0.005-0.024 across metrics
(Table~\ref{tab:execution-variability}).
However, within-protocol F1 ranges exceeded 0.10 for
5/19 oligo-family, 9/20 transition, 8/20 state, and
12/20 typed-edge comparisons
(Figure~\ref{fig:supp-execution-variability}).
The largest observed range was 0.471 for typed-edge F1
in s3-ATAC. Thus, aggregate performance remained broadly
consistent despite variability in individual protocols.

This analysis was limited to GPT-Codex; Figure~\ref{fig:baseline-native}
and Figure~\ref{fig:baseline-transition} retain one selected execution per 
system and protocol.
Protocol-bootstrap intervals in the main results reflect
across-protocol uncertainty conditional on these outputs,
rather than execution variability.
The results therefore do not quantify variability for
other systems or for the knowledge and learning interventions,
including matched $C_0$-$C_{30}$ comparisons.

\section{Reconstruction without protocol source bundles}
\label{app:name-only-control}
Publicly available sequencing protocols and resources such as \texttt{scg\_lib\_structs}
may have appeared in LLM training data, 
we evaluated whether models could reconstruct sequencing libraries 
using only protocol names and versions, without access to source documents. 
Generic instructions, output schemas,
and local tools remained available, but source documents and retrieval
access were withheld. External network access was restricted except
for model-provider connections, and agent worked within an isolated docker image. 
This test assesses performance
without supplied evidence, rather than distinguishing memorization
from general learned knowledge or reasoning.

\paragraph{Task~1 API control.}
We compared the PDF-input API predictions from
Figure~\ref{fig:task1-api-agent} with one independent
name-only API execution per model and protocol. 
Interestingly, name prompting 
generally yielded increased sequence similarity for all models, although 
only Claude remained significant
after Bonferroni correction across three comparisons
($p_{\mathrm{adj}}=0.0215$).
Exact reconstructions decreased
from 4 to 2 for GPT, 5 to 3 for Claude, and 4 to 3 for Gemini,
out of 20 protocols per model.

\paragraph{Native-agent comparison (Tasks~2-3).}
For Codex, Claude Code, and Antigravity, we compared
one Name execution with the original Bundle execution used in
Figure~\ref{fig:baseline-native} ($n=1$ per condition). All outputs were scored against
the same approved references.
All 20 protocols had valid pairs, with ddSEQ excluded from Task~2
($n=19$).
For each system and metric, we calculated the mean paired difference
$\Delta=\frac{1}{n}\sum_p(F_{p,\mathrm{Name}}-F_{p,\mathrm{Bundle}})$.
We used two-sided paired sign-flip tests with Bonferroni correction
across 12 comparisons. In contrast to the Task~1 similarity results,
Name prompting yielded lower mean F1 across all
systems and metrics, although none of the differences
remained significant after correction
(all $p_{\mathrm{adj}}>0.05$).

\input{tables/name_only_combined_tasks1-3.tex}

\clearpage
\section{Complementary model-harness performance metrics}
\begin{figure}[H]
\centering
\captionsetup[subfigure]{skip=1pt}

\begin{subfigure}[t]{\textwidth}
  \centering
  \includegraphics[height=0.225\textheight,keepaspectratio]
    {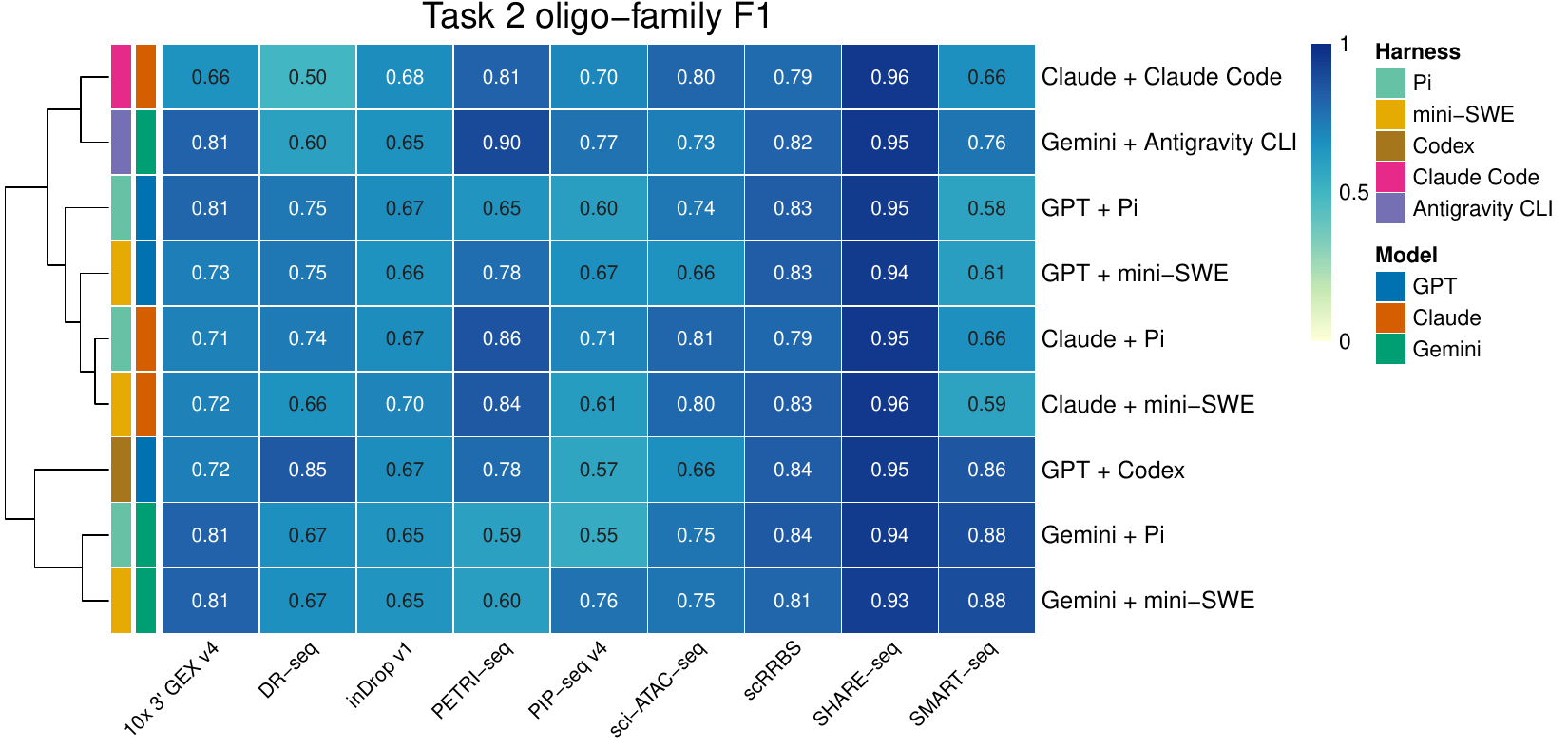}
  \caption{}
  \label{fig:supp-t2-heatmap}
\end{subfigure}

\vspace{0.2em}

\begin{subfigure}[t]{\textwidth}
  \centering
  \includegraphics[height=0.225\textheight,keepaspectratio]
    {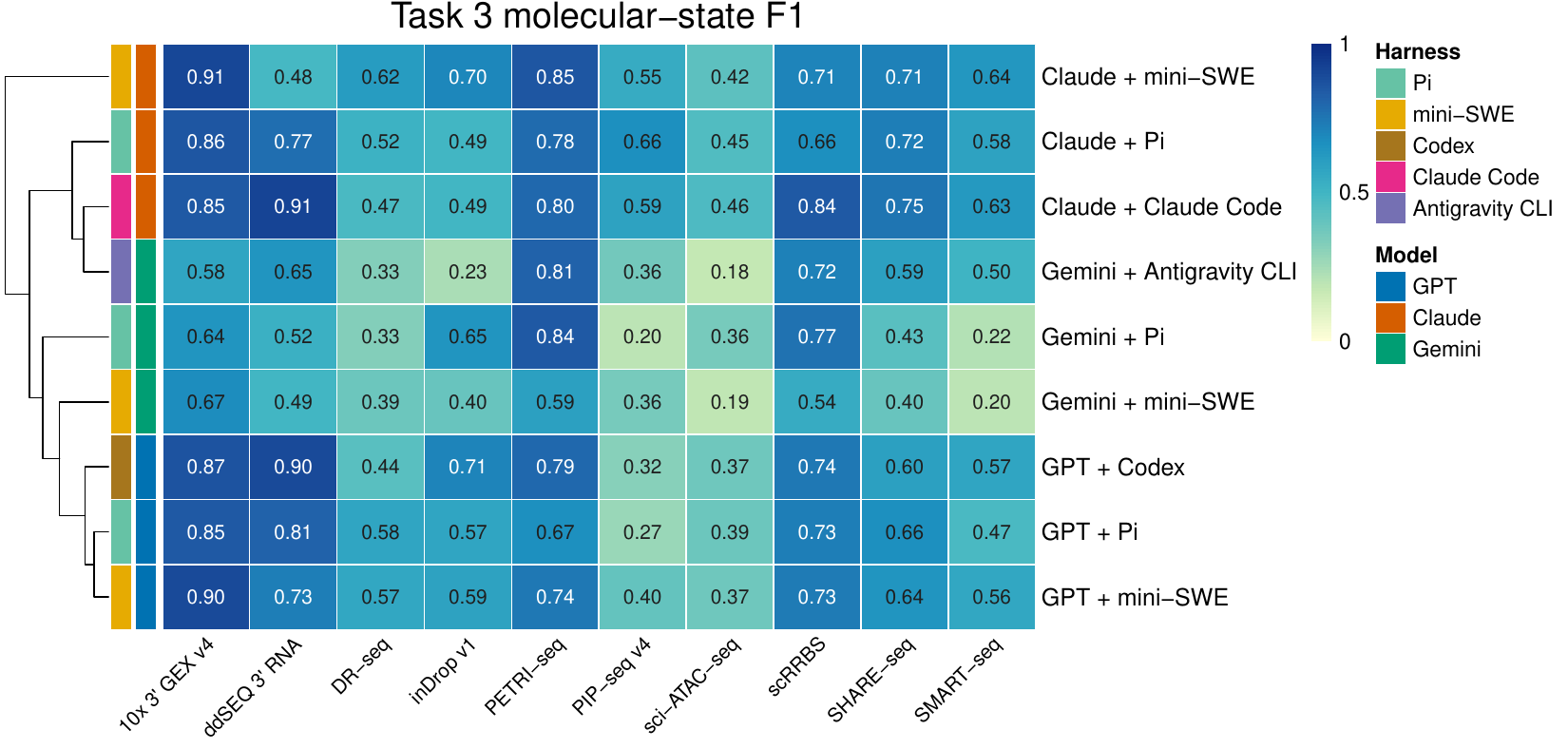}
  \caption{}
  \label{fig:supp-t3-state-heatmap}
\end{subfigure}

\vspace{0.2em}

\begin{subfigure}[t]{\textwidth}
  \centering
  \includegraphics[height=0.225\textheight,keepaspectratio]
    {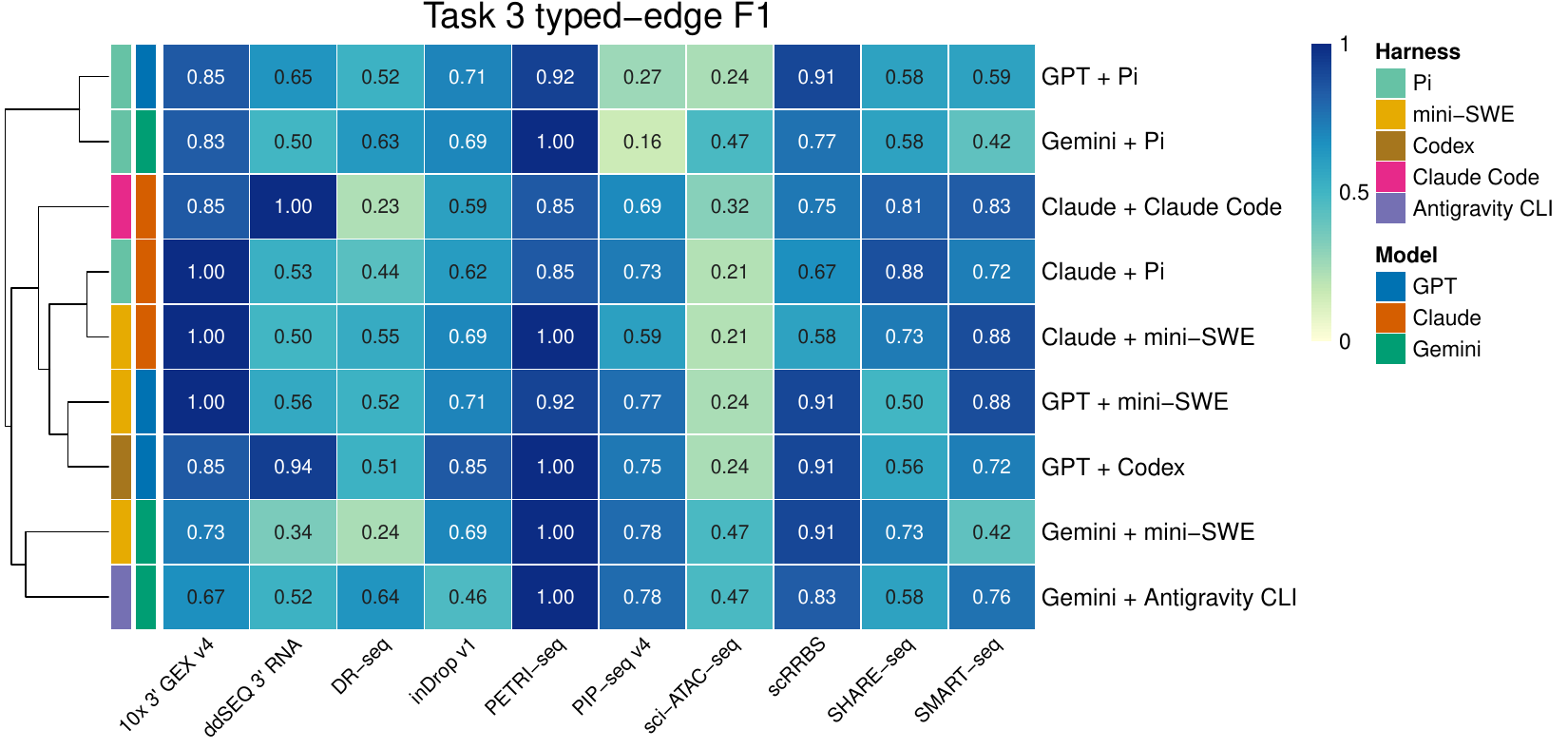}
  \caption{}
  \label{fig:supp-t3-edge-heatmap}
\end{subfigure}

\caption{%
  \textbf{Complementary performance metrics across models and harnesses.}
  Heatmaps show performance across the ten protocols and nine
  model--harness systems used in
  Figure~\ref{fig:baseline-transition}.
  System and protocol orders are fixed to the molecular-transition-F1
  clustering in the main figure to permit direct comparison across
  metrics. All color scales range from 0 to 1.%
}
\label{fig:supp-complementary-heatmaps}
\end{figure}

\clearpage
\section{Cost analysis and model-harness effects}
\begingroup
\setlength{\intextsep}{5pt}

\begin{figure}[h]
\centering
\captionsetup{font=small,skip=3pt}
\captionsetup[subfigure]{skip=1pt}
\begin{subfigure}[t]{0.48\textwidth}
  \centering
  \includegraphics[width=\linewidth,height=0.21\textheight,keepaspectratio]
    {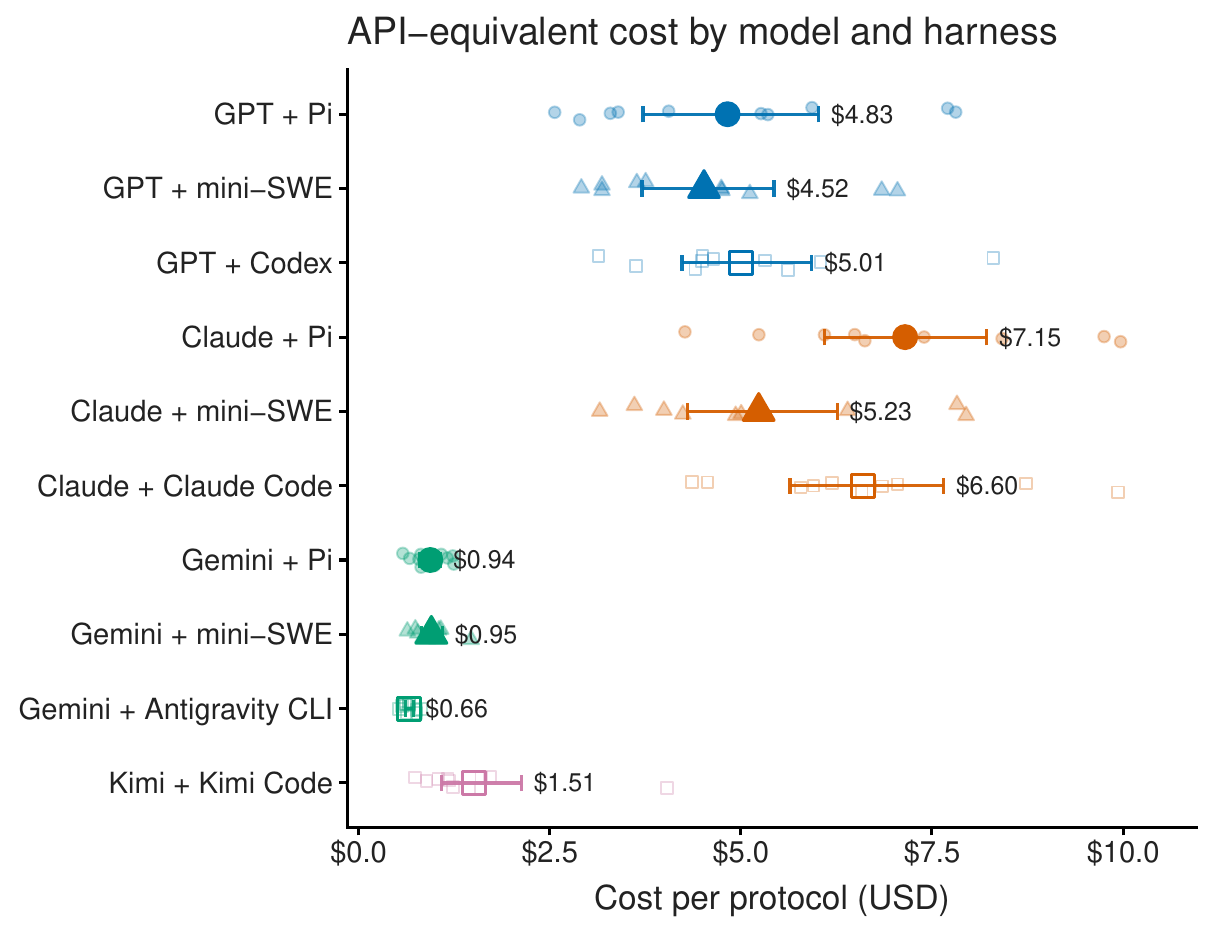}
  \caption{}\label{fig:supp-cost-by-system}
\end{subfigure}\hfill
\begin{subfigure}[t]{0.48\textwidth}
  \centering
  \includegraphics[width=\linewidth,height=0.21\textheight,keepaspectratio]
    {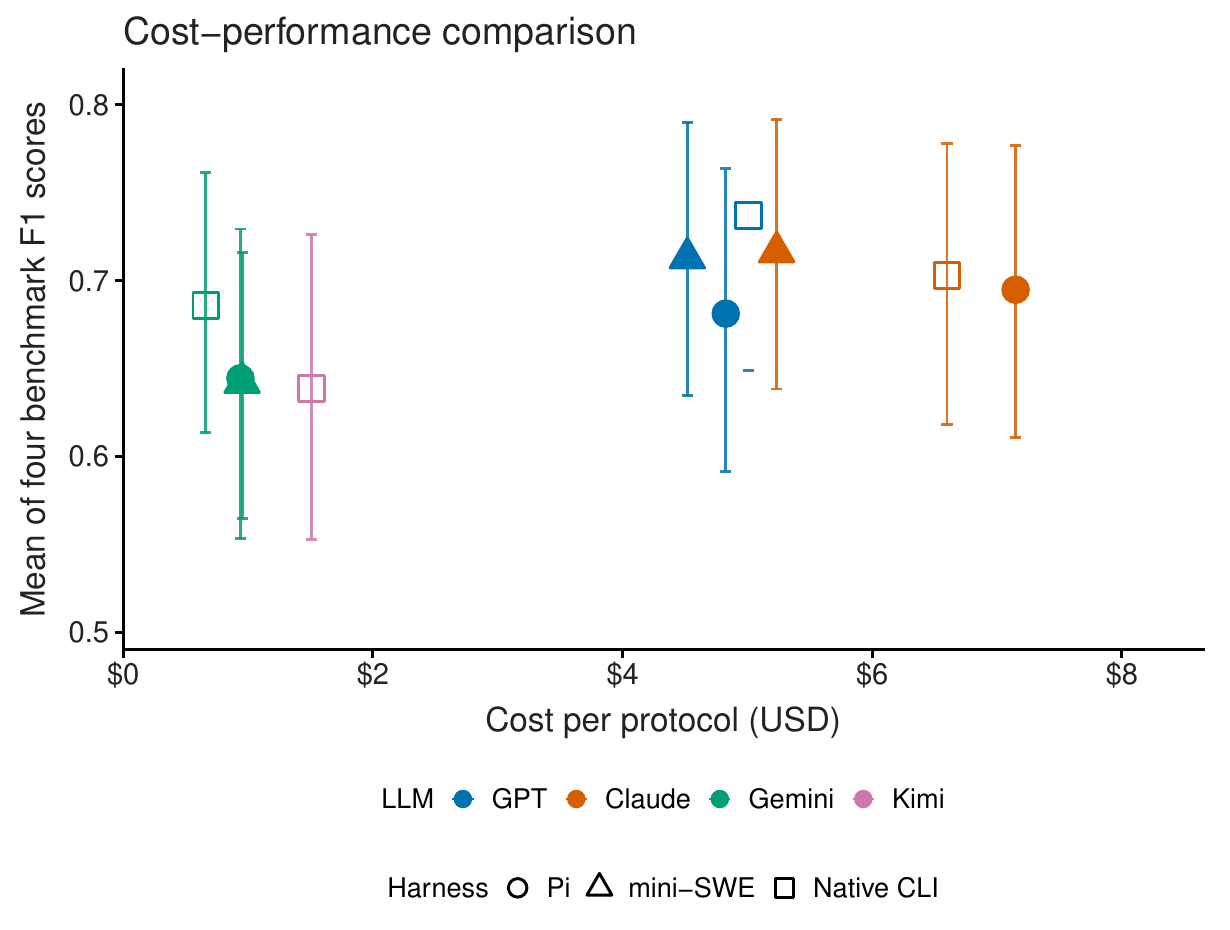}
  \caption{}\label{fig:supp-cost-performance}
\end{subfigure}
\caption{%
\textbf{Estimated API-equivalent cost and reconstruction performance.}
\textbf{(a)}~Per-protocol costs and system means across ten protocols.
\textbf{(b)}~Cost versus mean performance across four F1 metrics.%
}
\label{fig:supp-cost-analysis}
\end{figure}

\paragraph{Model-harness effect analysis.}
To test the differences of model and harness selection on reconstruction performance,
we compared GPT, Claude, and Gemini in a fully crossed design with
Pi and mini-SWE-agent, using one execution per configuration on 10
common protocols (excluding ddSEQ for Task~2) . For protocol $p$, model $m$, and harness $h$,
let $y_{pmh}$ denote F1. Model- and harness-switch magnitudes were
$M_p=\frac{1}{6}\sum_h\sum_{m<m'}|y_{pmh}-y_{pm'h}|$, $H_p=\frac{1}{3}\sum_m|y_{pm,\mathrm{mini}}-y_{pm,\mathrm{Pi}}|$.
The factors $1/6$ and $1/3$ average over the three unique
model pairs within each of two harnesses, and the single
harness pair within each of three models, respectively. 
We estimated the mean of $D_p=M_p-H_p$ and its pointwise 95\% CI
using 10,000 paired protocol-bootstrap resamples.  
Mean model-switch magnitudes exceeded harness-switch
magnitudes across all four metrics, but only molecular-state
F1 showed a significant difference
($\bar{D}=0.0793$, 95\% CI $[0.038,0.124]$,
$p_{\mathrm{Bonf.}}=0.0156$;
Figure~\ref{fig:supp-model-harness}).

\paragraph{Secondary analyses and results.}
For each metric, we fitted a linear mixed model with model, harness, and their interaction
as fixed effects and a protocol random intercept.
Protocol-blocked repeated-measures ANOVA with Greenhouse-Geisser 
correction served as a robustness check.
Bonferroni correction covered 12 omnibus tests (three effects
$\times$ four metrics) within each analysis.
Only molecular-state F1 showed a detectable model main effect
(mixed model: $F(2,45)=23.105$, $p_{\mathrm{adj}}=1.50\times10^{-6}$;
blocked analysis: $p_{\mathrm{adj}}=0.0010$).
State F1 averaged over the two harnesses was 0.655 for Claude,
0.612 for GPT, and 0.459 for Gemini. Neither analysis detected a
harness main effect or model-harness interaction for any metric.
These secondary tests concern average scores; nonsignificance
does not establish equivalence or negligible harness effects.

\begin{figure}[h]
\centering
\captionsetup{font=small,skip=3pt}
\includegraphics[width=0.78\textwidth]{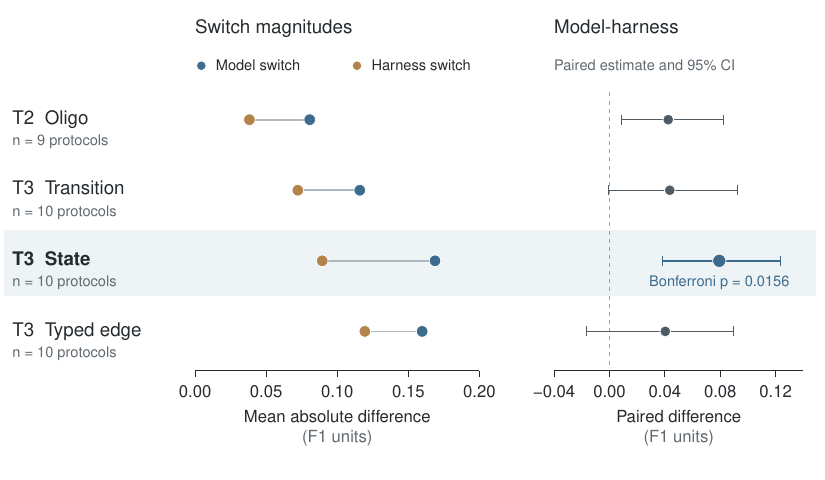}
\caption{%
\textbf{Model-harness effects.}
Mean absolute switch magnitudes (left) and paired model-minus-harness
contrasts (right). Intervals are pointwise 95\% protocol-bootstrap CIs,
conditional on the selected executions. The state annotation gives
$ p_{\mathrm{Bonf}} $ for the primary four-metric comparison.%
}
\label{fig:supp-model-harness}
\end{figure}
\endgroup

\clearpage
\section{Baseline output failure modes}
\begin{figure*}[h]
\centering
\captionsetup[subfigure]{skip=0pt}

\begin{subfigure}[t]{\textwidth}
  \centering
  \includegraphics[height=0.26\textheight,keepaspectratio]
    {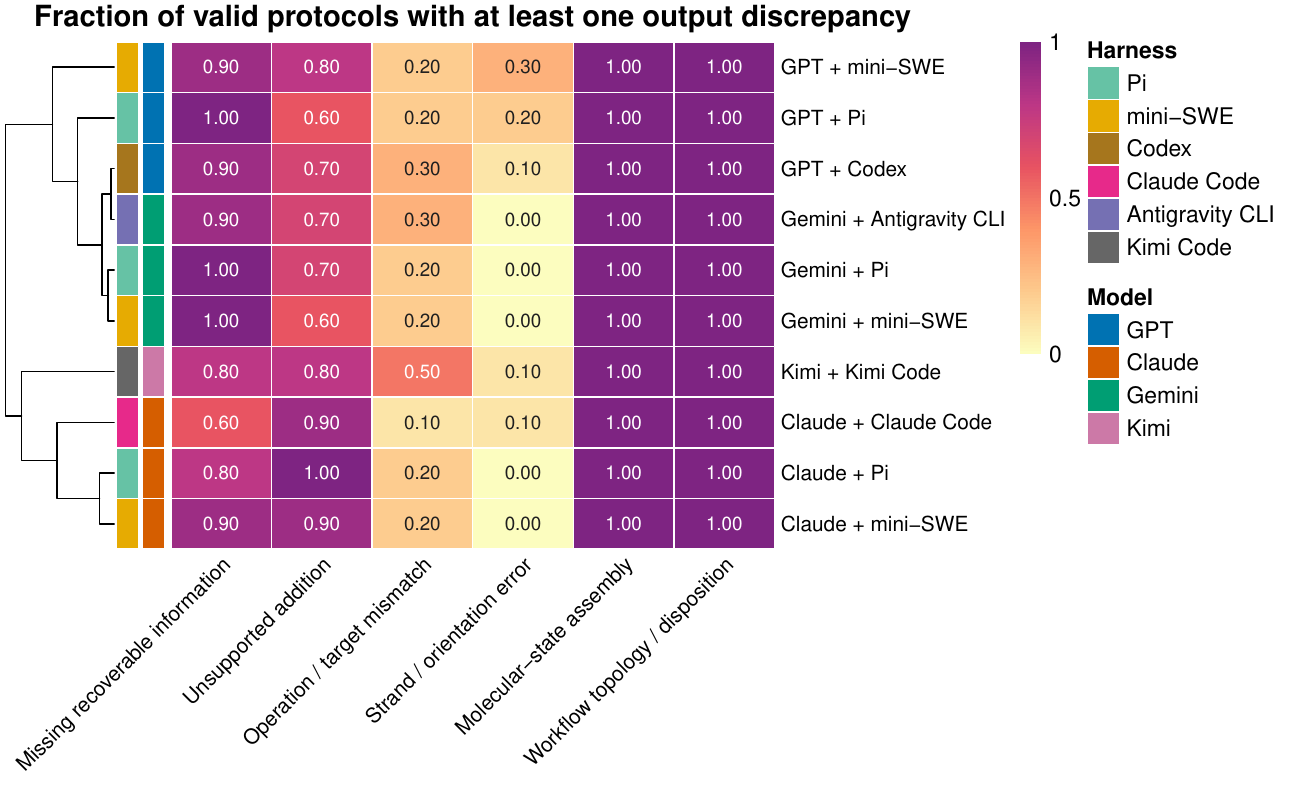}
  \caption{LLM-harness system profiles}
  \label{fig:supp-baseline-errors-system}
\end{subfigure}

\vspace{0.05em}

\begin{subfigure}[t]{\textwidth}
  \centering
  \includegraphics[height=0.26\textheight,keepaspectratio]
    {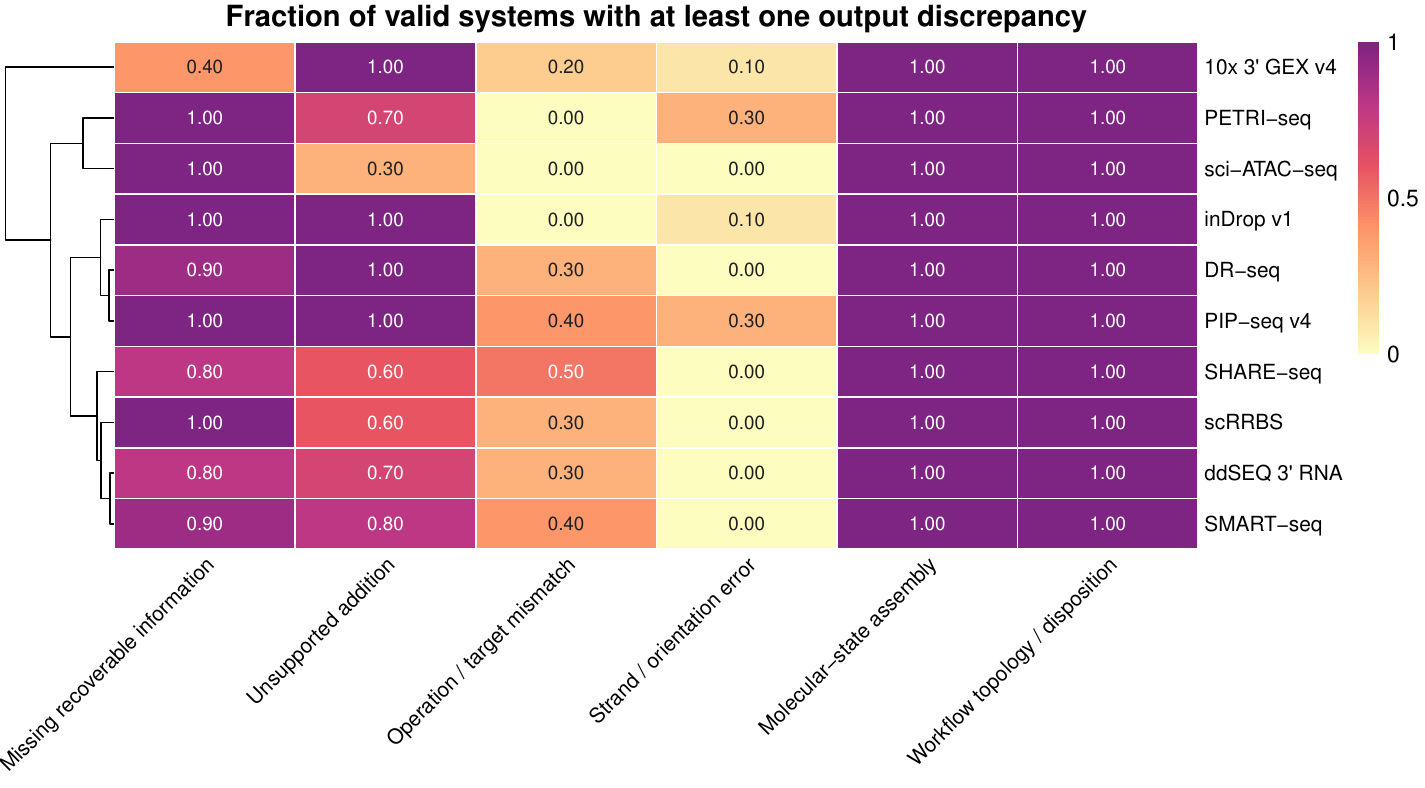}
  \caption{Protocol-centric profiles}
  \label{fig:supp-baseline-errors-protocol}
\end{subfigure}

\vspace{-0.1em}

\caption{%
\textbf{Baseline output-discrepancy profiles.}
\textbf{(a)}~The fraction of the ten protocols with at least one discrepancy.
\textbf{(b)}~The fraction of the ten systems with at least one discrepancy.
Values range from 0 to 1, where 1 indicates this error appeared in 100\% protocols
or LLM-harness runs.%
}
\label{fig:supp-baseline-errors}

\vspace{1.5em}

\begin{minipage}{0.94\textwidth}
\centering

\footnotesize
\setlength{\tabcolsep}{3pt}
\renewcommand{\arraystretch}{1.03}

\begin{tabularx}{\textwidth}{
  @{}
  >{\raggedright\arraybackslash}p{0.20\textwidth}
  >{\raggedright\arraybackslash}X
  >{\raggedright\arraybackslash}p{0.17\textwidth}
  @{}
}
\toprule
Error category & Description & Main effect \\
\midrule

Missing recoverable information &
Required source-supported content is missing or incomplete. &
Lower recall. \\

Unsupported addition &
The output adds content unsupported by the source and not marked optional or
neutral. &
Lower precision. \\

Operation / target mismatch &
An operation is incorrect or applied to the wrong molecular target. &
Lower transition F1. \\

Strand / orientation error &
A sequence or strand has an incorrect orientation or strand relationship. &
Lower oligo/state score. \\

Molecular-state assembly &
A state has incorrect strands, segment order, pairing, or architecture. &
Lower state F1. \\

Workflow topology / disposition &
A workflow connection or carried/discarded relationship is incorrect. &
Lower edge F1. \\

\bottomrule
\end{tabularx}
\captionof{table}{%
\textbf{Deterministic output-discrepancy categories used in the baseline
error analysis.}%
}
\label{tab:baseline-error-terms}
\end{minipage}
\end{figure*}

\clearpage
\section{External knowledge for agent improvement}

\begin{table*}[h]
\centering

\footnotesize
\setlength{\tabcolsep}{4pt}
\renewcommand{\arraystretch}{1.10}

\begin{tabularx}{\textwidth}{
@{}
>{\raggedright\arraybackslash}p{0.11\textwidth}
>{\raggedright\arraybackslash}p{0.48\textwidth}
>{\raggedright\arraybackslash}X
@{}
}
\toprule
Source type & Source & Primer content \\
\midrule

Textbook &
\emph{Molecular Cell Biology}, Lodish et al.\ 2003
\citep{lodish2003molecular} &
PCR, primer tails, restriction--ligation, reverse transcription, and DNA-end chemistry. \\

Textbook &
\emph{PCR Protocols}, Bartlett \& Stirling 2003
\citep{bartlett2003pcr} &
End repair, ligation, terminal modifications, and phosphate-dependent sealing. \\

Textbook &
\emph{Protocols for Gene Analysis}, Harwood 1994
\citep{harwood1994protocols} &
Primer tails, A-tailing, T-vector cloning, and blunt-end preparation. \\

Review &
\emph{Single-cell genome sequencing: current state of the science}, Gawad et al.\ 2016
\citep{gawad2016single} &
Whole-genome amplification, amplification artifacts, and library construction. \\

Review &
\emph{Scaling by shrinking: empowering single-cell `omics' with microfluidic devices}, Prakadan et al.\ 2017
\citep{prakadan2017scaling} &
Microfluidic compartmentalization, reaction splitting, and pooled processing. \\

Review &
\emph{Single-cell epigenomics: techniques and emerging applications}, Schwartzman \& Tanay 2015
\citep{schwartzman2015single} &
Bisulfite conversion, reduced-representation libraries, and size selection. \\

Review &
\emph{RNA sequencing: the teenage years}, Stark et al.\ 2019, 
\citep{stark2019rna} &
RNA-seq library preparation, reverse transcription, and amplification. \\

Review &
\emph{Methods and applications for single-cell and spatial multi-omics}, Vandereyken et al.\ 2023
\citep{vandereyken2023methods} &
Multimodal branching, combinatorial indexing, tagmentation, and template switching. \\

Review &
\emph{Chromatin accessibility profiling methods}, Minnoye et al.\ 2021
\citep{minnoye2021chromatin} &
Fragmentation, Tn5 tagmentation, adaptor addition, indexing, and amplification. \\

Research article &
\emph{Suppression of artifacts and barcode bias in high-throughput transcriptome analyses utilizing template switching}, Tang et al.\ 2013
\citep{tang2013suppression} &
Template switching, primer tails, barcode placement, and associated artifacts. \\

Review &
\emph{Time-resolved single-cell RNA-seq using metabolic RNA labelling}, Erhard et al.\ 2022 
\citep{erhard2022time} &
Affinity capture, reversible modification, elution, and molecular-fraction handling. \\

\bottomrule
\end{tabularx}
\caption{%
\textbf{Sources for the general molecular-method knowledge primer.}
The primer summarizes general molecular operations instead of protocol-specific answers.%
}
\label{tab:general-primer-sources}

\end{table*}

\begin{table*}[h]
\centering

\small
\setlength{\tabcolsep}{5pt}
\renewcommand{\arraystretch}{1.10}

\begin{tabularx}{\textwidth}{
@{}
>{\raggedright\arraybackslash}p{0.25\textwidth}
>{\raggedright\arraybackslash}p{0.25\textwidth}
>{\raggedright\arraybackslash}X
@{}
}
\toprule
Validation protocol & Donor protocol & Shared feature \\
\midrule

sci-ATAC-seq &
s3-ATAC &
Tn5-based chromatin accessibility and combinatorial indexing
\citep{cusanovich2015sciatac,mulqueen2021s3atac}. \\

scRRBS &
10x Chromium 3$'$ Gene Expression v4 &
Shared Illumina adapter and library-completion oligos
\citep{guo2015scrrbs,tenx2024gemxv4}. \\

SMART-seq &
Drop-seq &
Template-switching reverse transcription and cDNA amplification
\citep{ramskold2012smartseq,macosko2015dropseq}. \\

SHARE-seq &
SPLiT-seq &
Combinatorial indexing with repeated barcode-incorporation steps
\citep{ma2020shareseq,rosenberg2018splitseq}. \\

ddSEQ Single-Cell 3$'$ RNA-Seq Kit &
sci-RNA-seq &
Barcoded reverse transcription, cDNA amplification, and shared library
components \citep{biorad2024ddseq,cao2017scirnaseq}. \\

\bottomrule
\end{tabularx}
\caption{%
\textbf{Protocols used in the cross-protocol memory pilot.}
Donor protocols were selected as conceptual matches to the validation
protocols based on shared molecular features.%
}
\label{tab:pilot-protocol-pairing}
\end{table*}

\clearpage
\section{Trajectory-based behavioral changes across autonomous-learning checkpoints}
\label{app:checkpoint-behavior}

To interpret the non-monotonic learning curve, we compared the agent-visible
context, execution trajectories, and predictions at the baseline,
\(C_{30}\), \(C_{40}\), and \(C_{58}\) checkpoints on the same frozen
five-protocol validation panel. For each checkpoint, we analyzed one selected
execution per validation protocol. The observed changes support behavioral
interpretation but do not alone establish the causal effect of an individual
rule or retrieved example.

\begin{table*}[h]
\centering
\small
\setlength{\tabcolsep}{5pt}
\renewcommand{\arraystretch}{1.15}

\begin{tabularx}{\textwidth}{
@{}
>{\raggedright\arraybackslash}p{0.09\textwidth}
>{\raggedright\arraybackslash}p{0.18\textwidth}
>{\raggedright\arraybackslash}X
>{\raggedright\arraybackslash}p{0.27\textwidth}
@{}
}
\toprule
Checkpoint & Learned context & Observed reconstruction behavior &
Validation performance \\
\midrule

Baseline &
No learned rules or cross-protocol memory. &
Direct source-to-output reconstruction followed by the benchmark validator. &
Task~2 F1: 0.758. Task~3 transition, state, and typed-edge F1:
0.846, 0.625, and 0.822. \\

\(C_{30}\) &
18 learned rules and 30 available exemplar records. &
The agent constructed an evidence inventory, retrieved selected analogies,
and compiled Task~2 and Task~3 from a linked work record. The predicted
graphs averaged 10.0 states, 7.2 transitions, and 16.2 typed edges. &
Highest Task~2 F1 (0.821), with transition,
state, and typed-edge F1 of 0.843, 0.603, and 0.790.
Selected checkpoint for held-out evaluation. \\

\(C_{40}\) &
23 learned rules and 40 exemplar records. &
Two newly learned rules may have contributed to the observed decline.
A rule excluding oligo-reagent fabrication from the workflow was associated
with omitted bead-construction steps in HyDrop-RNA, while a rule
separating co-processed but physically unconnected molecules was associated
with the separation of the ddSEQ workflow into disconnected branches.
These behaviors suggest that learned rules may generalize poorly
to the benchmark's canonical representation. &
Task~2 F1 decreased to 0.736; transition, state, and typed-edge F1
decreased to 0.743, 0.497, and 0.662. \\

\(C_{58}\) &
28 learned rules and 58 exemplar records. &
The agent represented oligo chemistry, non-templated tail addition, modified
bases, and uncertain fragment fates more explicitly. Graphs again became more
granular. &
State F1 partially recovered to 0.539, while Task~2, transition,
and typed-edge F1 were 0.745, 0.720, and 0.638, respectively. \\

\bottomrule
\end{tabularx}

\caption{%
\textbf{Behavioral comparison of selected autonomous-learning checkpoints.}
Learned rules are protocol-neutral instructions accumulated from
previous training protocols; exemplar records are ground-truth-derived
cross-protocol examples available for retrieval.
Task~2 scores are macro-averaged over four informative protocols;
ddSEQ has no required recoverable oligo families and is reported
separately. Task~3 scores include all five validation protocols.%
}
\label{tab:checkpoint-behavior}
\end{table*}

The Task~3 deterioration visible at \(C_{40}\) was already apparent at
\(C_{35}\), with pronounced losses for ddSEQ and HyDrop-RNA.
The decline in informative Task~2 F1 also involved CEL-seq, whose
score decreased from 0.600 at \(C_{30}\) to 0.450 at \(C_{40}\).

For ddSEQ, a rule learned from a previous multi-omic protocol instructed
the agent to keep co-processed but physically distinct molecules in
separate workflows. In the observed trajectory, the agent separated the
cDNA and deconvolution-oligo branches into two workflows, whereas the
benchmark connects them through a shared fraction-separation transition.
One predicted workflow was therefore left unmatched.
From \(C_{30}\) to \(C_{40}\), ddSEQ transition F1 decreased
from 1.000 to 0.493, state F1 from 0.830 to 0.449, and
typed-edge F1 from 1.000 to 0.364.
Because ddSEQ has no required recoverable oligo families, it is excluded
from the Task~2 macro-average. Both checkpoints produced zero effective
spurious oligo families; \(C_{58}\) produced one.

For HyDrop-RNA, a learned rule instructed the agent to begin the molecular
workflow after oligo-reagent fabrication. In the observed reconstruction,
the agent represented a completed bead construct instead of the separate
Acrydite and split-pool barcode oligos, omitting early bead-construction
states and transitions. The benchmark, however, explicitly scores these
oligos and construction steps. From \(C_{30}\) to \(C_{40}\), HyDrop-RNA
Task~2 F1 decreased from 0.915 to 0.738, state F1 from 0.401 to 0.159,
transition F1 from 0.666 to 0.587, and typed-edge F1 from 0.609 to 0.526.
These cases are consistent with negative transfer or a mismatch between
learned representation conventions and the canonical benchmark.
The trajectories do not isolate the causal effect of an individual rule
from execution variability.

At \(C_{58}\), the agent recovered some molecular-state detail but frequently
used a finer event granularity than the canonical benchmark. For example,
it represented MspI-end fill-in, dA addition, bisulfite denaturation, and
bisulfite conversion as separate scRRBS events. This representation was
associated with higher scRRBS state F1, increasing from 0.521 at \(C_{40}\)
to 0.712 at \(C_{58}\), while transition F1 decreased from 0.883 to 0.741
and typed-edge F1 from 0.696 to 0.615, reflecting additional unmatched
transitions and edges. Similarly, the \(C_{58}\) chATAC-seq output
represented adapter reagents as independent molecular states, coinciding
with lower state precision and typed-edge agreement.

\section{Sensitivity analysis}
\label{app:metric-sensitivity}
Here we assessed the sensitivity of benchmark scores and system
comparisons to similarity weights and matching thresholds
used in Tasks~2 and~3.
In total, we evaluated 640 distinct scoring configurations:
one default configuration, three uniform-weight alternatives,
28 individual weight perturbations, 600 Dirichlet-sampled
weight vectors, and eight additional threshold combinations.
Weight and threshold variations were evaluated separately,
rather than through a full factorial combination.

For each configuration, we recomputed the complete matching
and scoring pipeline using the same saved agent predictions
and ground-truth records, without additional agent executions.
This captured changes in both partial-credit similarity
scores and molecular-state or transition alignments.
The resulting ranges characterize sensitivity to scoring
assumptions rather than statistical confidence intervals.

\paragraph{Sensitivity to similarity weights.}
We examined three sets of scientific similarity weights
for oligo-family, molecular-transition, and molecular-state
similarity. Table~\ref{tab:similarity-weights} summarizes
their default values and scientific rationale.
The weights were chosen heuristically to prioritize
molecular identity and relevant structural information,
rather than estimated from experimental data.

\begin{table}[h]
\centering
\small
\setlength{\tabcolsep}{4pt}
\renewcommand{\arraystretch}{1.15}

\begin{tabularx}{\textwidth}{
@{}
>{\raggedright\arraybackslash}p{0.16\textwidth}
>{\raggedright\arraybackslash}p{0.40\textwidth}
>{\raggedright\arraybackslash}X
@{}
}
\toprule
Metric & Default component weights & Scientific rationale \\
\midrule

Oligo F1 &
\textbf{Sequence: 0.65};
modifications: 0.15;
molecular kind: 0.10;
orientation: 0.05;
role: 0.05. &
Nucleotide sequence is central to oligo identity,
while modifications and molecular annotations
provide additional information. \\

Transition F1 &
\textbf{Operation: 0.30};
substrates: 0.15;
products: 0.20;
product disposition: 0.15;
linked oligos: 0.20. &
Emphasizes the molecular operation and its products
and linked oligos, while retaining information about
substrates and product disposition. \\

State F1 &
\textbf{Reference sequence: 0.50};
architecture: 0.15;
ordered segments: 0.20;
pairing/discontinuities: 0.15. &
Prioritizes molecular sequence identity while
accounting for strand architecture, segment
organization, and pairing relationships. \\

\bottomrule
\end{tabularx}
\vspace{0.5em}
\caption{%
\textbf{Default component weights for pairwise scientific similarity.}
These similarities contribute to entity matching and subsequent
F1 calculation. For oligo-family and molecular-state scoring,
weights are renormalized over components enabled by the
ground-truth support rules.%
}
\label{tab:similarity-weights}
\end{table}

For each default weight vector $\boldsymbol{w}_0$
(Table~\ref{tab:similarity-weights}),
we evaluated three types of alternatives:
i) uniform weights ($1/d$ for each of $d$ components);
ii) one-at-a-time $\pm20\%$ relative perturbations of
individual components, followed by renormalization;
and iii) 200 joint perturbations of all components,
sampled from
$\boldsymbol{w} \sim
\operatorname{Dirichlet}(\lambda\boldsymbol{w}_0)$,
with $\lambda=100$.
Here, $\lambda$ controls the dispersion of sampled weights
around the default vector $\boldsymbol{w}_0$.
The concentration parameter was chosen heuristically
to explore moderate departures from the default weights.
Each of the oligo, transition, and state weight vectors
was varied separately, with all other weights and
matching thresholds held fixed.

\begin{figure}[h]
\centering
\includegraphics[
  width=0.95\textwidth,
  keepaspectratio
]{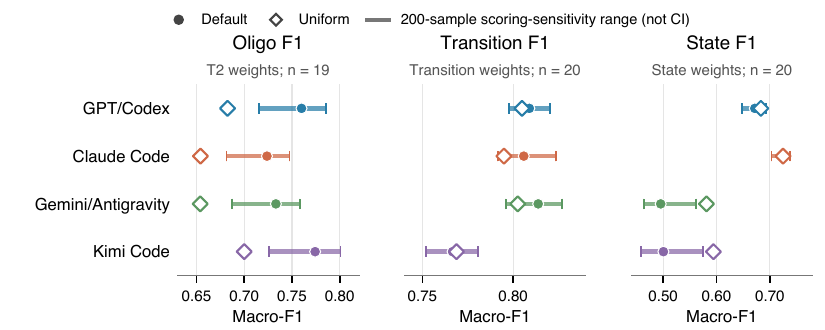}
\caption{%
\textbf{Sensitivity of native-system performance to similarity weights.}
Macro-averaged oligo-family, molecular-transition, and
molecular-state F1 scores for four native coding-agent systems.
The displayed ranges characterize sensitivity to scoring
assumptions, not statistical confidence intervals.
}
\label{fig:supp-sensitivity-weights}
\end{figure}

Figure~\ref{fig:supp-sensitivity-weights} shows that absolute
F1 scores varied with component weighting.
For example, uniform molecular-state weights increased
Gemini's state F1 from 0.495 to 0.581 and Kimi's from
0.500 to 0.594, while Claude remained the highest-scoring
system for molecular-state reconstruction.
The leading systems for oligo-family and molecular-state
F1 remained unchanged across the tested weight configurations.
Molecular-transition rankings were less stable: Gemini
ranked first in 179 of 200 sampled transition-weight
configurations, while GPT ranked first in the remaining 21.
Uniform transition weights also favored GPT.
These results indicate that the principal leading-system
comparisons were largely robust to alternative similarity
weights, although absolute F1 scores and the close
Gemini-GPT transition comparison were sensitive to
component weighting. The analysis supports the
stability of our comparative findings without establishing
that the default weights are biologically optimal.

\paragraph{Sensitivity to matching thresholds.}
Next we assessed sensitivity to molecular-state and transition
assignment thresholds, both set to 0.25 by default.
Each threshold was independently varied over
$\{0.15, 0.25, 0.40\}$, yielding nine combinations.
Thresholds were applied to context-adjusted assignment
scores rather than raw scientific similarity scores.
All similarity weights were kept at their default values,
and matching and scoring were recomputed for each combination.

Table~\ref{tab:threshold-sensitivity-summary} summarizes
the maximum absolute changes in macro-averaged F1 across
the tested thresholds and systems or conditions within
each experiment.
For the native-system comparison (Figure~\ref{fig:baseline-native}),
the maximum changes in transition, state, and typed-edge F1
were 0.035, 0.031, and 0.069, respectively.

\begin{table}[htbp]
\centering
\small
\setlength{\tabcolsep}{5pt}
\renewcommand{\arraystretch}{1.05}

\begin{tabular}{@{}lrrr@{}}
\toprule
Experiment & Transition F1 & State F1 & Typed-edge F1 \\
\midrule
Fig.~3a: Native systems              & 0.035 & 0.031 & 0.069 \\
Fig.~3b: Model-harness comparison    & 0.023 & 0.020 & 0.045 \\
Fig.~4a: External knowledge          & 0.029 & 0.013 & 0.054 \\
Fig.~4c: Learning validation         & 0.055 & 0.025 & 0.132 \\
Fig.~4d: Held-out evaluation         & 0.018 & 0.009 & 0.031 \\
\bottomrule
\end{tabular}

\vspace{0.5em}
\captionsetup{width=0.78\textwidth}

\caption{%
\textbf{Sensitivity to state and transition assignment thresholds.}
Maximum absolute change in protocol-macro F1 relative to
the default thresholds $(0.25, 0.25)$, across nine
threshold combinations and all systems or conditions
within each experiment.
}
\label{tab:threshold-sensitivity-summary}
\end{table}

\begin{figure}[h]
\centering
\includegraphics[
  width=0.95\textwidth,
  keepaspectratio
]{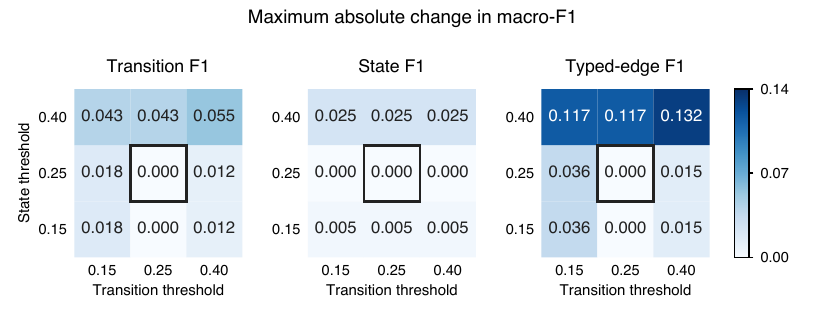}
\caption{%
\textbf{Sensitivity of molecular-workflow scores to matching thresholds.}
Heatmaps show the maximum absolute changes in molecular-transition,
molecular-state, and typed-edge F1 across autonomous-learning
validation checkpoints relative to their respective scores
under the default thresholds.
State and transition assignment thresholds were independently
varied over $\{0.15, 0.25, 0.40\}$.
Each cell represents the maximum across checkpoints,
rather than the average change.
The default combination $(0.25, 0.25)$ is outlined.
}
\label{fig:supp-sensitivity-thresholds}
\end{figure}

Figure~\ref{fig:supp-sensitivity-thresholds} illustrates
threshold sensitivity in the autonomous-learning validation
experiment (Figure~\ref{fig:agent-learning-curve}),
which exhibited the largest changes in transition and
typed-edge F1 (0.055 and 0.132, respectively).
Notably, increasing only the state-matching threshold
from 0.25 to 0.40 produced a maximum typed-edge F1
change of 0.117, compared with 0.025 for state F1.

Overall, transition and state F1 were less sensitive
to the tested matching thresholds than typed-edge F1.
This likely reflects the dependence of graph-connectivity
scoring on molecular-state alignment, where changes in
matched states can propagate to the evaluation of associated
edges. These findings suggest that graph-topology scores
should be interpreted alongside molecular-state alignment
and the canonical graph representation.

\begin{figure}[h]
\centering
\includegraphics[
  width=0.78\textwidth,
  height=0.23\textheight,
  keepaspectratio
]{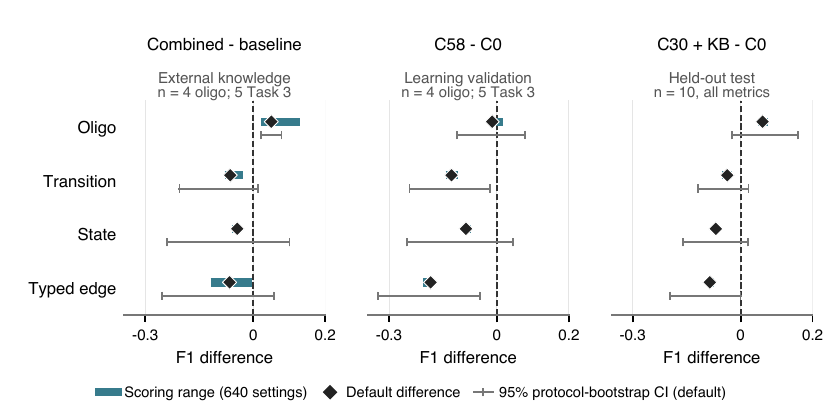}

\caption{%
\textbf{Robustness of knowledge augmentation and autonomous-learning results.}
Differences in F1 relative to baseline for
combined external knowledge versus baseline;
final checkpoint $C_{58}$ versus $C_0$
on the learning-validation panel; and 
$C_{30}$ with knowledge-base supplementation
versus $C_0$ on held-out protocols.
}
\label{fig:supp-sensitivity-interventions}
\end{figure}

\paragraph{Robustness of knowledge augmentation and cumulative learning.}
We further investigated whether the conclusions from the
external-knowledge and autonomous-learning experiments
were preserved under alternative scoring assumptions
(Figure~\ref{fig:supp-sensitivity-interventions}).
In the external-knowledge pilot
(Figure~\ref{fig:knowledge-pilot}), combining general
knowledge with cross-protocol examples consistently
increased oligo-family F1 relative to baseline
(range: 0.022--0.129) across the tested scoring
configurations. However, baseline retained the highest
mean scores across all three Task~3 metrics.
In the cumulative-learning validation experiment
(Figure~\ref{fig:agent-learning-curve}), performance
remained non-monotonic across all tested configurations.
The final-checkpoint oligo-family F1 difference
($C_{58}-C_0$) ranged from $-0.021$ to $0.015$,
indicating sensitivity to component weighting.
In contrast, all three Task~3 endpoint differences
remained negative.
On the held-out protocols
(Figure~\ref{fig:autonomous-test}), supplementing
the selected $C_{30}$ checkpoint with the knowledge
base produced consistently positive oligo-family F1
differences relative to baseline (range: 0.049--0.075).
However, the pointwise 95\% protocol-bootstrap
confidence interval under default scoring included
zero ([-0.024, 0.159]).
Both evaluated $C_{30}$ conditions remained below
baseline in mean transition, state, and typed-edge
F1 across the tested scoring configurations.

Together, these results indicate that the absence
of improvement in canonical molecular-workflow
reconstruction was robust to the tested scoring
assumptions. However, the final-checkpoint oligo-family
F1 comparison was sensitive to component weighting.

\paragraph{Limitations.}
These sensitivity analyses used fixed agent predictions
and therefore assess robustness to scoring assumptions,
rather than execution-to-execution variability
(Appendix~\ref{app:execution-variability}).
Robustness across the tested configurations does not
establish that the default weights are biologically
optimal or that canonical graph matching accommodates
all chemically valid representations.
Furthermore, the model-versus-harness statistical tests
were not repeated across all scoring configurations.

\end{document}

%% file: tables/name_only_combined_tasks1-3.tex
\begingroup
\fontsize{8pt}{9.0pt}\selectfont
\setlength{\tabcolsep}{1.65pt}
\renewcommand{\arraystretch}{1.10}
\setlength{\LTcapwidth}{\linewidth}
\begin{longtable}{@{}>{\raggedright\arraybackslash}p{0.235\linewidth}l*{10}{r}@{}}

\toprule
 & & \multicolumn{2}{c}{\textbf{Task~1}} & \multicolumn{2}{c}{\textbf{Task~2}} & \multicolumn{6}{c}{\textbf{Task~3}} \\
\cmidrule(lr){3-4}\cmidrule(lr){5-6}\cmidrule(l){7-12}
 & & \multicolumn{2}{c}{Similarity} & \multicolumn{2}{c}{Oligo F1} & \multicolumn{2}{c}{Transition F1} & \multicolumn{2}{c}{State F1} & \multicolumn{2}{c}{Edge F1} \\
\cmidrule(lr){3-4}\cmidrule(lr){5-6}\cmidrule(lr){7-8}\cmidrule(lr){9-10}\cmidrule(l){11-12}
Protocol & Model & PDF & Name & Bundle & Name & Bundle & Name & Bundle & Name & Bundle & Name \\
\midrule
\endfirsthead
\multicolumn{12}{@{}l}{\textit{Table~\thetable\ (continued)}}\\[2pt]
\toprule
 & & \multicolumn{2}{c}{\textbf{Task~1}} & \multicolumn{2}{c}{\textbf{Task~2}} & \multicolumn{6}{c}{\textbf{Task~3}} \\
\cmidrule(lr){3-4}\cmidrule(lr){5-6}\cmidrule(l){7-12}
 & & \multicolumn{2}{c}{Similarity} & \multicolumn{2}{c}{Oligo F1} & \multicolumn{2}{c}{Transition F1} & \multicolumn{2}{c}{State F1} & \multicolumn{2}{c}{Edge F1} \\
\cmidrule(lr){3-4}\cmidrule(lr){5-6}\cmidrule(lr){7-8}\cmidrule(lr){9-10}\cmidrule(l){11-12}
Protocol & Model & PDF & Name & Bundle & Name & Bundle & Name & Bundle & Name & Bundle & Name \\
\midrule
\endhead
\midrule
\multicolumn{12}{r@{}}{\textit{Continued on next page}}\\
\endfoot

\bottomrule
\noalign{\vskip 0.5em}
\caption{%
\textbf{Protocol-name-only reconstruction across Tasks~1--3.}
Task~1 reports API sequence similarity (PDF vs.\ Name);
Tasks~2--3 report native-agent F1 (Bundle vs.\ Name),
using the Figure~\ref{fig:baseline-native} bundle executions.
Bold indicates the higher score within each pair;
ties are unbolded.
$\Delta$ denotes Name minus source-input score.
$p_{\mathrm{adj}}$ denotes Bonferroni-adjusted $p$-values,
correcting separately across three Task~1 and
12 Tasks~2--3 comparisons.
}
\label{tab:name-only-all-tasks}\\
\endlastfoot
\textbf{10x ATAC v2} & GPT & \textbf{1.000} & 0.947 & \textbf{0.791} & 0.673 & 0.782 & \textbf{0.848} & \textbf{0.875} & 0.863 & 0.714 & \textbf{1.000} \\*
 & Claude & 1.000 & 1.000 & \textbf{0.713} & 0.687 & \textbf{0.777} & 0.761 & 0.776 & \textbf{0.812} & 0.714 & 0.714 \\*
 & Gemini & 1.000 & 1.000 & \textbf{0.701} & 0.673 & \textbf{0.956} & 0.707 & \textbf{0.720} & 0.396 & \textbf{1.000} & 0.667 \\
\addlinespace[1.7pt]
\textbf{10x 3' GEX v4} & GPT & 1.000 & 1.000 & 0.724 & \textbf{0.867} & 0.881 & \textbf{0.950} & 0.868 & \textbf{0.874} & 0.846 & \textbf{1.000} \\*
 & Claude & \textbf{1.000} & 0.970 & 0.658 & \textbf{0.849} & 0.881 & \textbf{0.999} & 0.847 & \textbf{0.875} & 0.846 & \textbf{1.000} \\*
 & Gemini & 1.000 & 1.000 & 0.811 & \textbf{0.845} & 0.858 & \textbf{0.869} & \textbf{0.578} & 0.462 & 0.667 & \textbf{0.818} \\
\addlinespace[1.7pt]
\textbf{10x 3' Feature Barcoding} & GPT & \textbf{1.000} & 0.853 & \textbf{0.865} & 0.703 & 0.695 & \textbf{0.751} & \textbf{0.720} & 0.550 & \textbf{0.650} & 0.600 \\*
 & Claude & \textbf{1.000} & 0.881 & \textbf{0.852} & 0.704 & \textbf{0.770} & 0.752 & \textbf{0.717} & 0.666 & 0.650 & 0.650 \\*
 & Gemini & \textbf{0.995} & 0.903 & \textbf{0.817} & 0.712 & \textbf{0.759} & 0.650 & \textbf{0.513} & 0.469 & 0.684 & 0.684 \\
\addlinespace[1.7pt]
\textbf{Drop-seq} & GPT & 0.897 & \textbf{0.989} & 0.675 & \textbf{0.733} & 0.766 & \textbf{0.918} & 0.534 & \textbf{0.739} & 0.750 & \textbf{0.960} \\*
 & Claude & 0.886 & \textbf{0.989} & 0.528 & \textbf{0.746} & 0.853 & \textbf{0.942} & 0.770 & \textbf{0.800} & 0.889 & \textbf{1.000} \\*
 & Gemini & 0.886 & \textbf{0.989} & 0.511 & \textbf{0.735} & \textbf{0.820} & 0.617 & \textbf{0.558} & 0.329 & \textbf{0.750} & 0.364 \\
\addlinespace[1.7pt]
\textbf{Seq-Well S3} & GPT & 0.681 & \textbf{1.000} & \textbf{0.834} & 0.733 & \textbf{0.829} & 0.676 & \textbf{0.801} & 0.643 & \textbf{0.778} & 0.643 \\*
 & Claude & 0.681 & \textbf{1.000} & \textbf{0.767} & 0.723 & 0.850 & \textbf{0.909} & \textbf{0.805} & 0.750 & 0.824 & \textbf{0.848} \\*
 & Gemini & 1.000 & 1.000 & \textbf{0.855} & 0.698 & 0.564 & \textbf{0.874} & 0.321 & \textbf{0.511} & 0.133 & \textbf{0.848} \\
\addlinespace[1.7pt]
\textbf{inDrop v1} & GPT & \textbf{1.000} & 0.586 & \textbf{0.672} & 0.448 & \textbf{0.895} & 0.766 & \textbf{0.706} & 0.505 & \textbf{0.846} & 0.720 \\*
 & Claude & \textbf{1.000} & 0.871 & \textbf{0.675} & 0.477 & 0.641 & \textbf{0.656} & 0.491 & \textbf{0.512} & \textbf{0.595} & 0.500 \\*
 & Gemini & \textbf{1.000} & 0.860 & \textbf{0.655} & 0.490 & 0.721 & \textbf{0.778} & 0.231 & \textbf{0.479} & 0.462 & \textbf{0.692} \\
\addlinespace[1.7pt]
\textbf{s3-ATAC} & GPT & 0.104 & \textbf{0.768} & \textbf{0.927} & 0.632 & \textbf{0.938} & 0.489 & \textbf{0.909} & 0.529 & \textbf{0.941} & 0.400 \\*
 & Claude & 0.098 & \textbf{0.793} & \textbf{0.897} & 0.618 & \textbf{0.988} & 0.486 & \textbf{0.840} & 0.553 & \textbf{0.941} & 0.400 \\*
 & Gemini & 0.000 & \textbf{0.817} & \textbf{0.884} & 0.611 & \textbf{0.986} & 0.488 & \textbf{0.474} & 0.239 & \textbf{0.941} & 0.400 \\
\addlinespace[1.7pt]
\textbf{CEL-seq} & GPT & 0.391 & \textbf{0.938} & 0.600 & \textbf{0.658} & 0.878 & \textbf{0.896} & 0.544 & \textbf{0.673} & 0.815 & \textbf{0.846} \\*
 & Claude & 0.391 & \textbf{0.993} & \textbf{0.600} & 0.518 & \textbf{0.900} & 0.876 & 0.718 & \textbf{0.751} & 0.815 & \textbf{0.846} \\*
 & Gemini & 0.384 & \textbf{0.968} & 0.353 & \textbf{0.683} & \textbf{0.835} & 0.832 & 0.315 & \textbf{0.333} & 0.692 & 0.692 \\
\addlinespace[1.7pt]
\textbf{SPLiT-seq} & GPT & 0.026 & \textbf{0.627} & \textbf{0.897} & 0.589 & \textbf{0.900} & 0.692 & \textbf{0.636} & 0.462 & \textbf{0.867} & 0.643 \\*
 & Claude & 0.132 & \textbf{0.756} & \textbf{0.901} & 0.731 & \textbf{0.771} & 0.752 & \textbf{0.816} & 0.791 & 0.714 & \textbf{0.857} \\*
 & Gemini & 0.170 & \textbf{0.667} & \textbf{0.878} & 0.546 & \textbf{0.800} & 0.785 & 0.406 & \textbf{0.417} & 0.714 & \textbf{0.846} \\
\addlinespace[1.7pt]
\textbf{sci-RNA-seq} & GPT & \textbf{0.900} & 0.874 & \textbf{0.771} & 0.754 & 0.664 & \textbf{0.824} & 0.825 & \textbf{0.829} & 0.636 & \textbf{0.778} \\*
 & Claude & 1.000 & 1.000 & \textbf{0.797} & 0.730 & 0.698 & \textbf{0.825} & \textbf{0.818} & 0.727 & 0.636 & \textbf{0.778} \\*
 & Gemini & \textbf{0.900} & 0.542 & 0.743 & \textbf{0.760} & 0.811 & \textbf{0.824} & \textbf{0.466} & 0.461 & 0.778 & 0.778 \\
\addlinespace[1.7pt]
\textbf{sci-ATAC-seq} & GPT & 0.049 & \textbf{0.663} & 0.661 & \textbf{0.675} & 0.564 & \textbf{0.677} & \textbf{0.373} & 0.329 & 0.235 & \textbf{0.471} \\*
 & Claude & 0.098 & \textbf{0.741} & \textbf{0.804} & 0.647 & \textbf{0.694} & 0.492 & \textbf{0.463} & 0.402 & \textbf{0.316} & 0.211 \\*
 & Gemini & 0.000 & \textbf{0.741} & \textbf{0.729} & 0.662 & \textbf{0.694} & 0.590 & 0.179 & \textbf{0.188} & \textbf{0.471} & 0.235 \\
\addlinespace[1.7pt]
\textbf{Microwell-seq} & GPT & 0.578 & \textbf{0.826} & 0.527 & \textbf{0.739} & 0.900 & \textbf{0.944} & 0.648 & \textbf{0.714} & 0.909 & \textbf{1.000} \\*
 & Claude & 0.578 & \textbf{0.853} & \textbf{0.516} & 0.475 & \textbf{0.808} & 0.544 & \textbf{0.732} & 0.240 & \textbf{0.783} & 0.182 \\*
 & Gemini & 0.509 & \textbf{0.812} & 0.545 & \textbf{0.661} & \textbf{0.940} & 0.666 & \textbf{0.734} & 0.362 & \textbf{1.000} & 0.600 \\
\addlinespace[1.7pt]
\textbf{scRRBS} & GPT & 0.953 & 0.953 & \textbf{0.838} & 0.710 & \textbf{0.949} & 0.926 & \textbf{0.741} & 0.645 & 0.909 & 0.909 \\*
 & Claude & 0.953 & 0.953 & \textbf{0.790} & 0.418 & 0.813 & \textbf{0.925} & \textbf{0.845} & 0.754 & 0.750 & \textbf{0.909} \\*
 & Gemini & 0.945 & \textbf{0.953} & \textbf{0.816} & 0.528 & 0.809 & \textbf{0.915} & \textbf{0.718} & 0.371 & 0.833 & \textbf{0.909} \\
\addlinespace[1.7pt]
\textbf{PIP-seq v4} & GPT & 0.829 & \textbf{0.842} & 0.575 & \textbf{0.713} & 0.765 & \textbf{0.888} & 0.316 & \textbf{0.637} & 0.750 & \textbf{0.960} \\*
 & Claude & 0.829 & \textbf{0.838} & 0.697 & \textbf{0.731} & \textbf{0.771} & 0.747 & 0.589 & \textbf{0.658} & 0.686 & \textbf{0.720} \\*
 & Gemini & \textbf{0.829} & 0.820 & \textbf{0.767} & 0.635 & \textbf{0.775} & 0.682 & 0.363 & \textbf{0.382} & \textbf{0.783} & 0.720 \\
\addlinespace[1.7pt]
\textbf{scDamID} & GPT & \textbf{0.994} & 0.595 & \textbf{0.643} & 0.496 & \textbf{0.783} & 0.547 & \textbf{0.644} & 0.546 & \textbf{0.692} & 0.480 \\*
 & Claude & \textbf{0.994} & 0.637 & 0.629 & \textbf{0.728} & \textbf{0.881} & 0.789 & \textbf{0.672} & 0.627 & \textbf{0.846} & 0.750 \\*
 & Gemini & \textbf{0.994} & 0.619 & \textbf{0.661} & 0.607 & \textbf{0.992} & 0.490 & \textbf{0.452} & 0.255 & \textbf{1.000} & 0.364 \\
\addlinespace[1.7pt]
\textbf{ddSEQ} & GPT & 0.063 & \textbf{0.401} & \multicolumn{1}{c}{--} & \multicolumn{1}{c}{--} & \textbf{0.936} & 0.701 & \textbf{0.898} & 0.669 & \textbf{0.938} & 0.519 \\*
 & Claude & 0.181 & \textbf{0.382} & \multicolumn{1}{c}{--} & \multicolumn{1}{c}{--} & \textbf{0.971} & 0.700 & \textbf{0.914} & 0.649 & \textbf{1.000} & 0.583 \\*
 & Gemini & 0.063 & \textbf{0.384} & \multicolumn{1}{c}{--} & \multicolumn{1}{c}{--} & 0.664 & \textbf{0.700} & \textbf{0.648} & 0.490 & 0.519 & \textbf{0.583} \\
\addlinespace[1.7pt]
\textbf{SMART-seq} & GPT & 0.000 & \textbf{0.342} & \textbf{0.859} & 0.811 & \textbf{0.791} & 0.604 & \textbf{0.573} & 0.499 & \textbf{0.722} & 0.643 \\*
 & Claude & 0.240 & \textbf{0.500} & 0.662 & \textbf{0.833} & \textbf{0.879} & 0.800 & 0.630 & \textbf{0.631} & \textbf{0.833} & 0.800 \\*
 & Gemini & 0.000 & \textbf{0.500} & 0.757 & \textbf{0.866} & \textbf{0.803} & 0.557 & \textbf{0.501} & 0.276 & \textbf{0.765} & 0.462 \\
\addlinespace[1.7pt]
\textbf{SHARE-seq} & GPT & 0.083 & \textbf{0.457} & \textbf{0.948} & 0.542 & \textbf{0.760} & 0.674 & \textbf{0.596} & 0.399 & \textbf{0.562} & 0.536 \\*
 & Claude & 0.083 & \textbf{0.770} & \textbf{0.955} & 0.653 & \textbf{0.821} & 0.789 & \textbf{0.754} & 0.598 & \textbf{0.812} & 0.714 \\*
 & Gemini & 0.167 & \textbf{0.609} & \textbf{0.950} & 0.534 & \textbf{0.784} & 0.349 & \textbf{0.591} & 0.243 & \textbf{0.583} & 0.231 \\
\addlinespace[1.7pt]
\textbf{DR-seq} & GPT & 0.109 & \textbf{0.630} & \textbf{0.850} & 0.425 & \textbf{0.565} & 0.323 & 0.435 & \textbf{0.465} & \textbf{0.511} & 0.300 \\*
 & Claude & 0.109 & \textbf{0.771} & \textbf{0.497} & 0.338 & \textbf{0.482} & 0.422 & 0.472 & \textbf{0.484} & \textbf{0.226} & 0.213 \\*
 & Gemini & 0.109 & \textbf{0.882} & \textbf{0.600} & 0.444 & \textbf{0.722} & 0.483 & 0.329 & \textbf{0.336} & \textbf{0.643} & 0.426 \\
\addlinespace[1.7pt]
\textbf{PETRI-seq} & GPT & 0.568 & \textbf{0.668} & \textbf{0.782} & 0.424 & \textbf{0.939} & 0.765 & \textbf{0.793} & 0.638 & \textbf{1.000} & 0.818 \\*
 & Claude & 0.563 & \textbf{0.569} & \textbf{0.814} & 0.525 & \textbf{0.869} & 0.687 & \textbf{0.796} & 0.589 & \textbf{0.846} & 0.667 \\*
 & Gemini & \textbf{0.985} & 0.596 & \textbf{0.897} & 0.092 & \textbf{0.983} & 0.844 & \textbf{0.811} & 0.523 & 1.000 & 1.000 \\
\midrule
\textbf{Macro mean} & GPT & 0.561 & \textbf{0.748} & \textbf{0.760} & 0.649 & \textbf{0.809} & 0.743 & \textbf{0.672} & 0.610 & \textbf{0.754} & 0.711 \\*
 & Claude & 0.591 & \textbf{0.813} & \textbf{0.724} & 0.638 & \textbf{0.806} & 0.743 & \textbf{0.723} & 0.644 & \textbf{0.736} & 0.667 \\*
 & Gemini & 0.597 & \textbf{0.783} & \textbf{0.733} & 0.620 & \textbf{0.814} & 0.685 & \textbf{0.495} & 0.376 & \textbf{0.721} & 0.616 \\
\addlinespace[2pt]
\textbf{$\Delta$ (Name$-$source)} & GPT & \multicolumn{2}{c}{$+0.187$} & \multicolumn{2}{c}{$-0.111$} & \multicolumn{2}{c}{$-0.066$} & \multicolumn{2}{c}{$-0.061$} & \multicolumn{2}{c}{$-0.042$} \\*
 & Claude & \multicolumn{2}{c}{$+0.223$} & \multicolumn{2}{c}{$-0.085$} & \multicolumn{2}{c}{$-0.063$} & \multicolumn{2}{c}{$-0.080$} & \multicolumn{2}{c}{$-0.069$} \\*
 & Gemini & \multicolumn{2}{c}{$+0.186$} & \multicolumn{2}{c}{$-0.113$} & \multicolumn{2}{c}{$-0.129$} & \multicolumn{2}{c}{$-0.119$} & \multicolumn{2}{c}{$-0.105$} \\
\addlinespace[2pt]
\textbf{$p_{\mathrm{adj}}$} & GPT & \multicolumn{2}{c}{0.0566} & \multicolumn{2}{c}{0.2535} & \multicolumn{2}{c}{1.0000} & \multicolumn{2}{c}{1.0000} & \multicolumn{2}{c}{1.0000} \\*
  & Claude & \multicolumn{2}{c}{0.0215} & \multicolumn{2}{c}{0.5449} & \multicolumn{2}{c}{0.9889} & \multicolumn{2}{c}{0.1331} & \multicolumn{2}{c}{1.0000} \\*
  & Gemini & \multicolumn{2}{c}{0.1329} & \multicolumn{2}{c}{0.7261} & \multicolumn{2}{c}{0.1420} & \multicolumn{2}{c}{0.1123} & \multicolumn{2}{c}{1.0000} \\
\addlinespace[2pt]
\textbf{Task~1 exact} & GPT & \textbf{4/20} & 2/20 & \multicolumn{8}{c}{} \\*
 & Claude & \textbf{5/20} & 3/20 & \multicolumn{8}{c}{} \\*
 & Gemini & \textbf{4/20} & 3/20 & \multicolumn{8}{c}{} \\
\end{longtable}

\endgroup